\documentclass[lettersize,journal]{IEEEtran}
\usepackage{amsmath,amsfonts}
\usepackage{algorithmic}
\usepackage{algorithm}
\usepackage{array}
\usepackage[caption=false,font=normalsize,labelfont=sf,textfont=sf]{subfig}
\usepackage{textcomp}
\usepackage{stfloats}
\usepackage{url}
\usepackage{verbatim}
\usepackage{graphicx}
\usepackage{cite}
\usepackage[utf8]{inputenc} % allow utf-8 input
\usepackage[T1]{fontenc}    % use 8-bit T1 fonts
\usepackage{hyperref}       % hyperlinks
\usepackage{booktabs}       % professional-quality tables
\usepackage{nicefrac}       % compact symbols for 1/2, etc.
\usepackage{microtype}      % microtypography
\usepackage{wrapfig, lipsum}
\usepackage[dvipsnames]{xcolor}
\usepackage{pifont}
\usepackage{enumitem}
\usepackage{color}
\usepackage{makecell}
\usepackage[table]{xcolor}

\usepackage{xcolor}
\newif\iftrackchanges
\trackchangesfalse % <-- OFF

\iftrackchanges
  \usepackage[normalem]{ulem}
  \newcommand{\addedfirst}[1]{\textcolor{blue}{#1}}
  \newcommand{\addedsecond}[1]{\textcolor{red}{#1}}
  \newcommand{\revcommentfirst}[1]{\textcolor{teal}{[\textbf{Changes for Revision 1: #1}]}~}
  \newcommand{\revcommentsecond}[1]{\textcolor{violet}{[\textbf{Changes for Revision 2: #1}]}~}
\else
  \newcommand{\addedfirst}[1]{#1}
  \newcommand{\addedsecond}[1]{#1}
  \newcommand{\revcommentfirst}[1]{}
  \newcommand{\revcommentsecond}[1]{}
\fi

\newcommand{\cmark}{\checkmark}
\newcommand{\xmark}{\scalebox{0.75}{\usym{2613}}}

\hypersetup{hidelinks}

\usepackage[many]{tcolorbox}
\usepackage{multirow}
\usepackage{utfsym}
\usepackage{rotating}
\usepackage{tablefootnote}

\usepackage{caption}
\definecolor{lightgreen}{HTML}{D5E8D4}
\definecolor{lightgray}{HTML}{D3D3D3}

\newcommand{\rot}[1]{\rotatebox[origin=l]{75}{#1}}

\begin{document}

\title{Chat-Scene++: Exploiting Context-Rich Object Identification for 3D LLM}
% \title{Chat-Scene++: Harnessing Context-Rich Object Sequences for 3D Scene LLMs}
% \title{Chat-Scene++: Integrating Object Identifiers and Context-Rich Representations for 3D LLMs}

\author{Haifeng Huang,
        % <-this % stops a space
        Yilun Chen,~\IEEEmembership{Member,~IEEE,}\\
        Zehan Wang,
        Jiangmiao Pang,
        Zhou Zhao,~\IEEEmembership{Member,~IEEE}
}% <-this % stops a space

\markboth{Journal of \LaTeX\ Class Files,~Vol.~14, No.~8, August~2021}%
{Shell \MakeLowercase{\textit{et al.}}: A Sample Article Using IEEEtran.cls for IEEE Journals}


\maketitle

\renewcommand{\thefootnote}{}
\footnotetext{\fontsize{7.8pt}{9pt}\selectfont Received 24 March 2025; revised 3 February 2026; accepted 28 March 2026. This work was supported in part by the National Natural Science Foundation of China under Grant U24A20326 and in part by the National Natural Science Foundation of China under Grant 62572423. Recommended for acceptance by M.-M. Cheng. \textit{(Haifeng Huang and Yilun Chen contributed equally to this work.) (Corresponding author: Zhou Zhao.)}}
\footnotetext{\fontsize{7.8pt}{9pt}\selectfont Haifeng Huang, Zehan Wang, and Zhou Zhao are with the School of Computer Science and Technology, Zhejiang University, Hangzhou 310027, China.}
\footnotetext{\fontsize{7.8pt}{9pt}\selectfont Yilun Chen and Jiangmiao Pang are with Shanghai AI Laboratory, Shanghai 200032, China.}
\footnotetext{\fontsize{7.8pt}{9pt}\selectfont Code will be made available at \href{https://github.com/Collab-Gen/Chat-Scene}{https://github.com/Collab-Gen/Chat-Scene}.}
\footnotetext{\fontsize{7.8pt}{9pt}\selectfont This article has supplementary downloadable material available at \href{https://doi.org/10.1109/TPAMI.2026.3679561}{https://doi.org/10.1109/TPAMI.2026.3679561}, provided by the authors.}
\footnotetext{\fontsize{7.8pt}{9pt}\selectfont Digital Object Identifier 10.1109/TPAMI.2026.3679561}
% \footnotetext{\textbullet~Work done during an internship at Shanghai AI Laboratory.}
% \footnotetext{$^*$ Equal contribution. $^\dag$ Corresponding author.}
% \footnotetext{© IEEE. This article has been accepted for publication in \textit{IEEE Transactions on Pattern Analysis and Machine Intelligence}. This is the author’s accepted manuscript and may change prior to final publication.}

\begin{abstract}  
Recent advancements in multi-modal large language models (MLLMs) have shown strong potential for 3D scene understanding. However, existing methods struggle with fine-grained object grounding and contextual reasoning, limiting their ability to interpret and interact with complex 3D environments. In this paper, we present Chat-Scene++, an MLLM framework that represents 3D scenes as context-rich object sequences. By structuring scenes as sequences of objects with contextual semantics, Chat-Scene++ enables object-centric representation and interaction. It decomposes a 3D scene into object representations paired with identifier tokens, allowing LLMs to follow instructions across diverse 3D vision-language tasks. To capture inter-object relationships and global semantics, Chat-Scene++ extracts \textit{context-rich} object features using large-scale pre-trained 3D scene-level and 2D image-level encoders, unlike the isolated per-object features in Chat-Scene. Its flexible object-centric design also supports grounded chain-of-thought (G-CoT) reasoning, enabling the model to distinguish objects at both category and spatial levels during multi-step inference. Without the need for additional task-specific heads or fine-tuning, Chat-Scene++ achieves state-of-the-art performance on five major 3D vision-language benchmarks: ScanRefer, Multi3DRefer, Scan2Cap, ScanQA, and SQA3D. These results highlight its effectiveness in scene comprehension, object grounding, and spatial reasoning. Additionally, without reconstructing 3D worlds through computationally expensive processes, we demonstrate its applicability to real-world scenarios using only 2D inputs.
\end{abstract}

\begin{IEEEkeywords}
Multi-modal Large Language Models, 3D Scene Understanding
\end{IEEEkeywords}

\section{Introduction}
\label{sec:intro}

Large Language Models (LLMs)~\cite{chiang2023vicuna, gpt4, touvron2023llama, chowdhery2022palm, ye2023mplug, li2023otter} have revolutionized the landscape of artificial intelligence, establishing language as a universal interface for building general-purpose assistants. This progress has catalyzed the development of Multi-modal LLMs (MLLMs), which integrate vision and language capabilities to address complex multi-modal tasks. Although 2D MLLMs~\cite{li2023videochat, llava, zhao2023bubogpt, zhu2023minigpt, llava1-5, lisa, ferret} achieve strong performance, extending these capabilities to 3D scene understanding remains challenging due to complex spatial relations, occlusions, and object interactions.

\begin{figure}[t]
    \centering
    \includegraphics[width=0.45\textwidth]{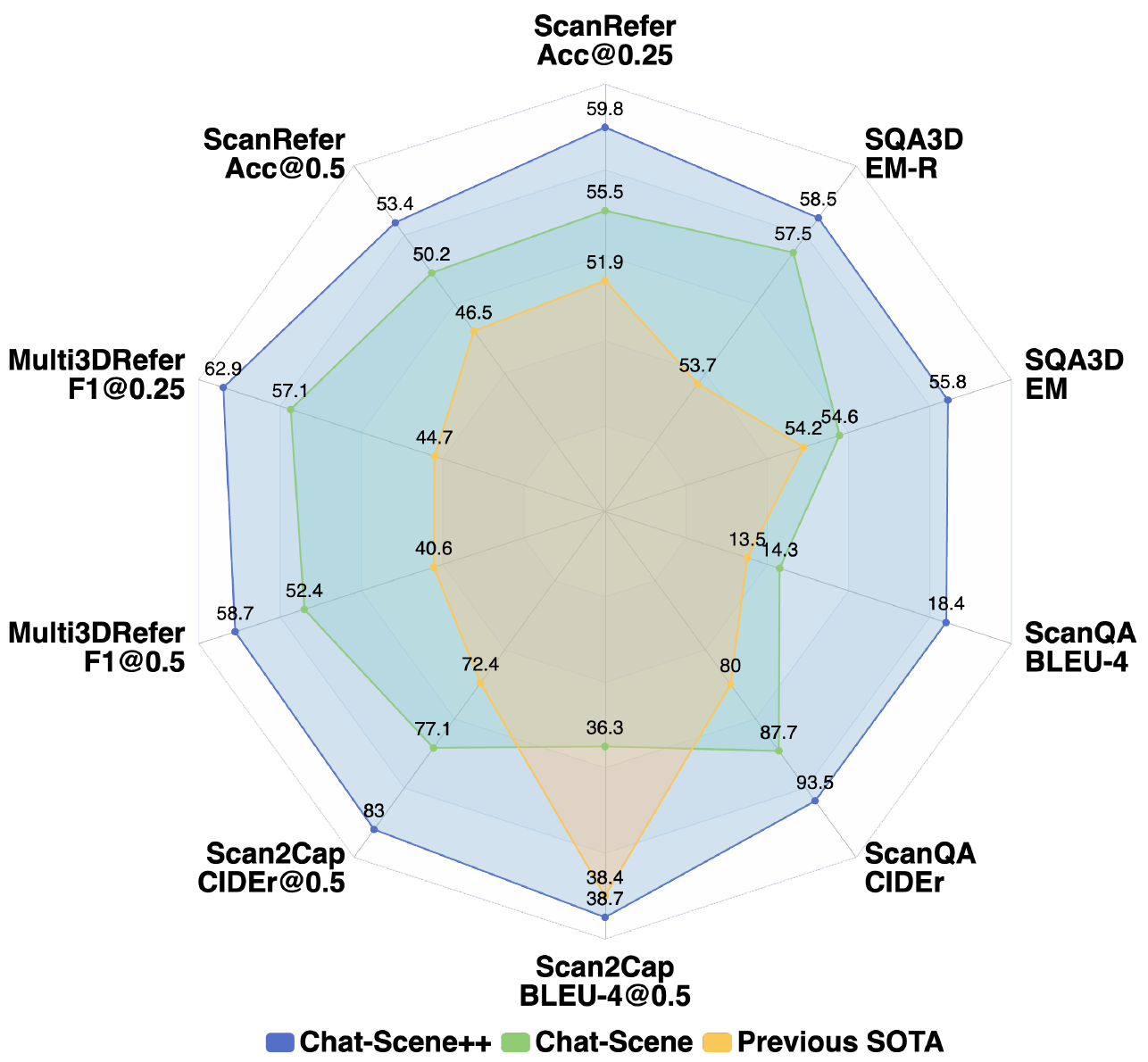}
    \caption{Chat-Scene++ surpasses previous SOTA methods on five benchmarks in the field of 3D scene understanding.}
    \label{fig:radar_chart}
\end{figure}

In 3D vision-language (VL) tasks such as 3D grounding~\cite{scanrefer, multi3drefer, referit3d}, 3D captioning~\cite{scan2cap}, and 3D/situated QA~\cite{scanqa, sqa3d}, models are evaluated for semantic understanding, spatial reasoning, object referencing, and grounding. Traditionally, these tasks relied on specialized methods: referencing requires interpreting a user-specified object, while grounding requires localizing it in the scene. Recent 3D MLLMs seek a unified 3D scene understanding model, yet many still lack robust referencing and grounding, especially when precise localization is required, limiting their general-purpose utility. For referencing, several 3D MLLMs~\cite{ll3da, leo, chat3d} add prompt encoders to parse user-specified objects but do not support grounding. \mbox{3D-LLM}~\cite{3dllm} introduces location tokens~\cite{ferret} to enable grounding; however, these approaches still lag behind expert models, particularly on grounding benchmarks. This weakness in referencing and grounding prevents 3D LLMs from reliably mapping 3D scenes into appropriate language representations.

To address these limitations, we propose \textbf{Chat-Scene++}, motivated by \textit{structuring 3D scenes as context-rich object sequences} to enable object-centric modeling and interaction. Specifically, we represent a scene as a sequence of object features paired with identifier tokens, facilitating flexible object referencing and relational reasoning during language interaction. This unified representation supports robust scene understanding and object-level reasoning across diverse 3D vision-language tasks.

\begin{figure}[t]
    \centering
    \includegraphics[width=0.85\linewidth]{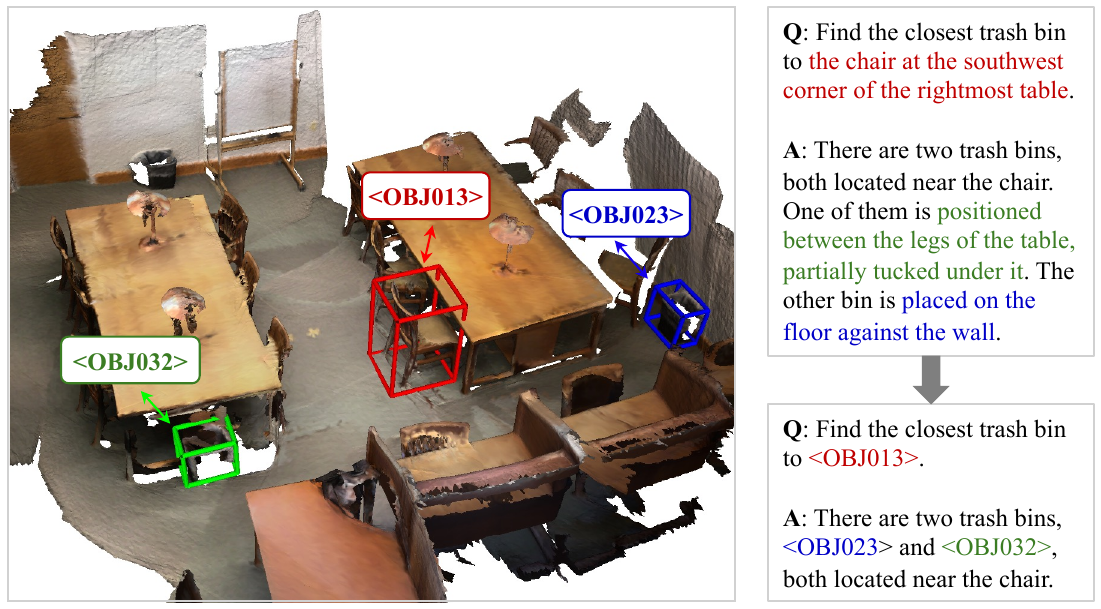}
    \caption{Example of utilizing object identifiers in conversation.}
    \label{fig:intro}
\end{figure}

\noindent\textit{\textbf{Reference 3D scene using object identifiers.}}
Objects define the semantic and spatial structure of a scene—their presence, location, and relationships shape the entire 3D context. This observation is central to many 3D scene understanding tasks such as grounding, VQA, and captioning, which all operate at the object level. We propose decomposing the 3D scene into a set of object proposals and assigning each object a unique identifier (ID)—a set of learnable tokens $\left \{\texttt{<OBJ}k\texttt{>} \right\}_{k=1...n}$—to distinguish them during language modeling. This design allows the LLM to reference respective objects using discrete ID tokens. As the example shown in Figure~\ref{fig:intro}, the chair and the two trash cans are labeled as ``$\texttt{<OBJ013>}$'', ``$\texttt{<OBJ023>}$'', and ``$\texttt{<OBJ032>}$'', respectively. This avoids the text ambiguity that arises from subjective viewing words like ``rightmost''. Moreover, verbose descriptions like “the chair located at the southwest corner of the rightmost table” can be replaced with concise, unambiguous identifiers, streamlining user-assistant interaction. This design allows us to convert various 3D VL tasks into unified question-answering formats without requiring task-specific heads or fine-tuning, facilitating scalable and efficient multi-task training.

\noindent\textit{\textbf{Represent 3D scenes using context-rich object sequences.}}  
Learning global scene-level representations often requires large-scale paired scene-language data, which is expensive and impractical for real-world environments. Instead, we represent scenes as sequences of context-rich object embeddings, extracted from both 3D and 2D modalities. This sequence-based modeling captures inter-object relations and global context without requiring dense scene-level supervision. In contrast to Chat-Scene~\cite{chatscene}, which used an independent object detector~\cite{mask3d} and separate encoder~\cite{uni3d} -- thus treating objects in isolation and causing redundant computation -- we introduce a unified, context-aware representation strategy. Chat-Scene++ employs a 3D scene-level encoder to generate object proposals and directly extract context-rich 3D features. These proposals are projected onto multi-view 2D images, from which a 2D encoder extracts complementary visual features, enhancing both spatial and semantic context. The fusion of multi-modal context-rich features, both pre-trained on large-scale data, captures fine-grained semantics and spatial layout. This forms a unified multi-modal representation of the scene, where each object is not only semantically grounded but also enriched with contextual cues. This sequential format enables the LLM to interpret complex scenes with object-level granularity and contextual awareness. Empirical results show that this structured representation substantially enhances the model’s performance on grounding and spatial reasoning tasks, validating the benefits of modeling scenes as context-aware object sequences.

% To enhance the contextual modeling of object-centric representations, we introduce a multi-modal approach that integrates both 3D and 2D features in a context-rich manner. The core intuition behind this method is that combining multiple perspectives—such as 3D object representations and 2D visual cues—provides a richer and more holistic understanding of the scene. For 3D feature extraction, we employ a 3D scene-level encoder to generate object proposals while capturing contextually rich features. For 2D feature extraction, we project these object proposals onto multi-view images and extract complementary features using a 2D image-level encoder, enhancing spatial and visual context. Leveraging large-scale pre-training, both 3D and 2D features encode deep semantic and structural relationships within the scene. These features are then mapped into the language model’s embedding space and combined with object IDs, forming a unified sequence of object-level embeddings. This results in scene-level representations enriched with multi-modal, context-aware information, ultimately enhancing the LLM’s performance, as demonstrated in our experiments—particularly on visual grounding tasks that demand complex scene-aware understanding and reasoning.  

\noindent\textit{\textbf{Enable multi-step object-level reasoning via grounded chain-of-thought.}}  
Modeling scenes as sequences of identifiable objects allows grounded, multi-step reasoning over both semantic and spatial relationships. While benchmarks like ScanQA~\cite{scanqa} and ScanRefer~\cite{scanrefer} typically provide short-form supervision (e.g., an answer token or bounding box), they often lack the intermediate reasoning steps required for complex scene understanding. For example, in a counting task, simply answering ``4'' does not specify which four objects were counted. Inspired by chain-of-thought prompting~\cite{cot_reasoning}, we introduce \textit{grounded chain-of-thought} (G-CoT), which explicitly integrates object IDs into multi-step reasoning. Unlike previous multi-modal CoT approaches~\cite{llava-cot, insight-v} that focus on textual reasoning, our method anchors reasoning steps to specific objects within the scene. Experiments demonstrate that grounded CoT not only improves performance on grounding and QA tasks requiring complex spatial reasoning but also enhances the interpretability of the model’s answers by providing intermediate reasoning steps tied to specific objects. This approach offers valuable insights for refining 3D LLMs and paves the way for more advanced intermediate reasoning capabilities in future models, similar to how GPT-o1 and related methods~\cite{openai-o1, guo2025deepseek} approach multi-step inference.

% \begin{figure}[t]
%     \centering
%     \includegraphics[width=0.49\textwidth]{figures/chat-scene-compare.pdf}
%     \caption{A comparison between our proposed Chat-Scene++ and its preliminary version, Chat-Scene. Chat-Scene++ replaces the object-level detect-and-encode pipeline with a unified scene-level encoder, which improves spatial relationship modeling by jointly extracting object masks and context-rich features. Additionally, Chat-Scene++ effectively integrates grounded chain-of-thought reasoning, further enhancing its ability to perform complex multi-step inference.}
%     \label{fig:chat-scene-compare}
% \end{figure}

We evaluate our approach on five major 3D scene-language benchmarks: ScanRefer~\cite{scanrefer}, Multi3DRefer~\cite{multi3drefer}, Scan2Cap~\cite{scan2cap}, ScanQA~\cite{scanqa}, and SQA3D~\cite{sqa3d}. As shown in Figure~\ref{fig:radar_chart}, our model consistently achieves state-of-the-art performance across all benchmarks, demonstrating its effectiveness in scene comprehension, object grounding, and spatial reasoning. Furthermore, without relying on computationally expensive 3D reconstruction, we showcase its applicability to real-world scenarios using only 2D inputs. Experiments on ScanNet~\cite{scannet} and VSI-Bench~\cite{vsi-bench} further validate its effectiveness for 2D-based visual grounding and VQA.

\section{Related Work}
\label{sec:formatting}

\revcommentsecond{We have revised the Related Work section for clarity and concision to meet the page limit.}\noindent\textbf{3D scene-language understanding}
Recent 3D scene understanding increasingly uses language to supply contextual knowledge and query conditions for precise intent interpretation, a paradigm often termed ``3D scene-language understanding.'' The primary tasks include:  1) 3D Visual Grounding~\cite{scanrefer, referit3d, tgnn, mvt, multi3drefer, vil3drel, 3drp-net, 3dvg-transformer, wang2023distilling, concretenet}, which localizes referred objects from text; 2) 3D Dense Captioning~\cite{scan2cap, X-trans2cap, more, vote2cap, vote2cap++}, which densely localizes and captions objects; 3) 3D Visual Question Answering~\cite{scanqa, parelli2023clip, sqa3d}, which answers general scene questions. Initial efforts concentrated on specialized tasks, resulting in limited generalizability across different 3D scene understanding tasks. Early methods were task-specific and generalized poorly. Recent work (e.g., 3DJCG~\cite{3djcg}, D3Net~\cite{d3net}) jointly models grounding and dense captioning to exploit synergy, while frameworks such as 3D-VisTA~\cite{3dvista} and 3D-VLP~\cite{3d-vlp} pursue broader scene-language alignment via pre-training. Despite improved multi-task coverage, these models still rely on task-specific heads, limiting their flexibility for general user-assistant interaction.

\noindent\textbf{Multi-modal large language models.}
Recent LLMs exhibit strong reasoning and human-like dialogue~\cite{chiang2023vicuna, gpt4, touvron2023llama, chowdhery2022palm, mllm-survey}, motivating extensions to additional modalities~\cite{lisa, li2023videochat, llava, zhao2023bubogpt, zhu2023minigpt, han2023imagebind, pointbind, 3dllm, chat3d, ye2023mplug, li2023otter, pointllm, leo, grounded3dllm}. In 3D, PointLLM~\cite{pointllm} maps point clouds into the LLM embedding space, while Imagebind-LLM~\cite{han2023imagebind} and Point-LLM~\cite{pointbind} learn a shared embedding across point clouds, images, audio, and text. Although effective on object-level tasks, these models struggle with complex 3D spatial relations. 3D-LLM~\cite{3dllm} adds positional embeddings and location tokens, but still projects 3D features through pre-trained 2D VLM input spaces, limiting 3D structural understanding. Chat-3D~\cite{chat3d} aligns 3D scene-text data directly with the LLM via pre-alignment to mitigate data scarcity, yet its architecture limits targeting specific objects in interaction. More recent 3D MLLMs~\cite{leo, ll3da, scenellm} improve QA and captioning but remain weak on visual grounding. Grounded 3D-LLM~\cite{grounded3dllm} introduces referent tokens, though its additional alignment objective yields suboptimal performance. By incorporating object identifiers into the LLM, our approach unifies object referencing and grounding for 3D MLLMs, demonstrating great potential for complex real-world applications.

% \noindent\textbf{3D Representation Learning} Recently, numerous efforts have been made to learn discriminative and robust representations for 3D point clouds, which serve as a fundamental visual modality. Approaches such as PointBERT~\cite{yu2022point}, Point-MAE~\cite{pang2022masked}, Transformer-OcCo~\cite{wang2021unsupervised}, and Point-m2ae~\cite{zhang2022point} employ self-supervised learning techniques to extract meaningful representations of 3D objects from unlabeled point cloud data. Another set of works~\cite{xue2023ulip, liu2023openshape, zhang2023learning, dong2022autoencoders, wang2023connecting, wang2023extending, freebind, omnibind} seeks to extend representation from other modalities to the 3D domain. For example, ULIP~\cite{xue2023ulip} and OpenShape~\cite{liu2023openshape} construct 3D-image-text triplets to align point clouds within the CLIP~\cite{radford2021learning} representation space. They leverage knowledge from existing MCR spaces to tackle to challenge of lacking paired data. These robust 3D representations effectively capture detailed information about 3D objects. Our approach involves segmenting the 3D scene at the instance level and extracting a set of object features to represent the entire scene.

\noindent\textbf{3D representation learning}
Effective 3D representation learning is key to scene understanding~\cite{mm-int, mm-rep}. Self-supervised methods (e.g., PointBERT~\cite{yu2022point}, Point-MAE~\cite{pang2022masked}, Point-m2ae~\cite{zhang2022point}) learn object-level features from raw point clouds via masked reconstruction and contrastive learning. Cross-modal approaches such as ULIP~\cite{xue2023ulip} and OpenShape~\cite{liu2023openshape} align 3D with vision-language models like CLIP~\cite{CLIP} using large-scale multi-modal training. More recently, Grounded 3D-LLM~\cite{grounded3dllm} proposed Contrastive Language-Scene Pre-training (CLASP) for phrase-level scene-text alignment at scale, supporting object proposal generation and semantically rich representations. We follow this pretraining strategy to extract context-rich features.

\begin{figure*}[t]
    \centering
    \includegraphics[width=0.95\textwidth]{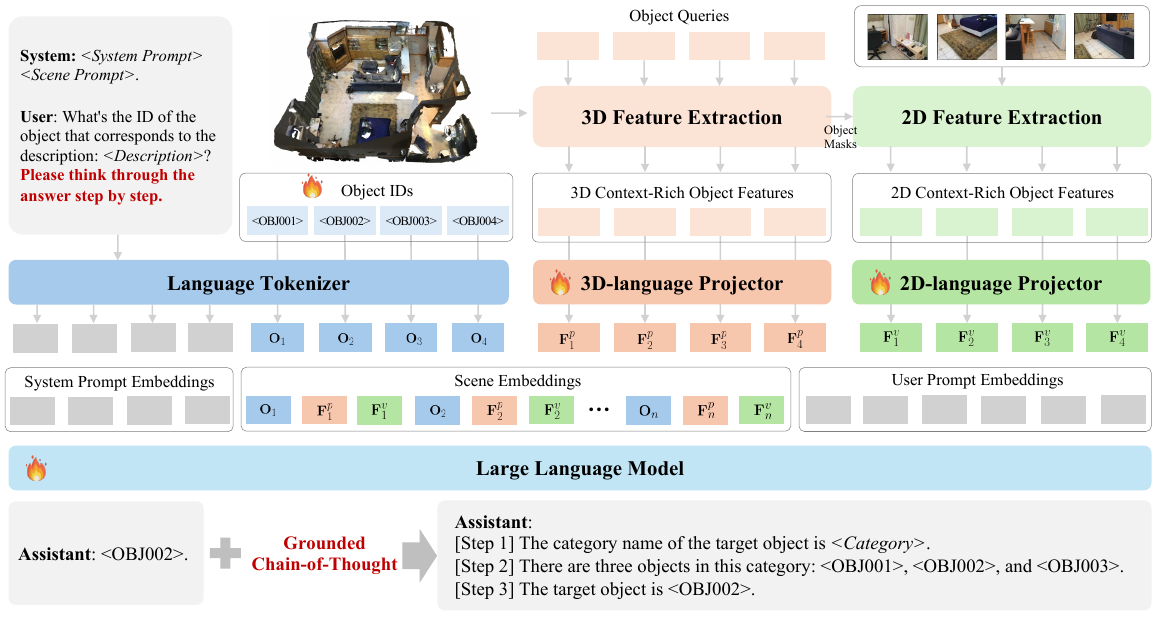}
    % \caption{\textbf{Overall model architecture of Chat-Scene++.} The model first extracts 3D features and generates object proposals from point clouds. Based on these proposals, 2D features are extracted from multi-view images. Both 2D and 3D features are mapped into the language model's token space. Object identifiers are incorporated into the vocabulary, linking to object proposals for efficient referencing and grounding. The final object embedding sequence represents the 3D scene as input to the LLM.}
    \caption{\textbf{Overall model architecture of Chat-Scene++.} The model structures a 3D scene as a context-rich object sequence, forming the scene embeddings for the LLM input. Specifically, it decomposes the 3D scene into a sequence of object representations paired with object ID tokens. Context-rich object features are extracted using large-scale pre-trained 3D and 2D models. These features are then mapped to the LLM's embedding space, where scene embeddings are constructed by sequentially combining object IDs with their corresponding object-level embeddings. By leveraging flexible object IDs, Grounded CoT can be optionally enabled to enhance reasoning over object relationships.}
    \label{fig:model}

\end{figure*}

\section{Method\label{sec:method}}

\revcommentsecond{We have revised the Method section for clarity and concision to meet the page limit.}This section presents \textbf{Chat-Scene++}, a unified framework for precise object referencing and context-rich 3D scene understanding in 3D MLLMs. We represent scenes as sequences of object-level embeddings, use object identifiers for explicit grounding, and leverage large-scale pre-trained models to capture contextual semantics. Section~\ref{sec:3d_scene_modeling} describes 3D scene input modeling; Section~\ref{sec:3d_llm_modeling} details integrating 3D scene information with LLMs; Section~\ref{sec:grounded_cot} introduces \textit{grounded chain-of-thought} for object-aware reasoning; and Section~\ref{sec:training_strategy} outlines the training strategy.

\subsection{3D Scene Modeling\label{sec:3d_scene_modeling}}
To enable object referencing and grounding in the LLM, Chat-Scene++ represents each 3D scene as a context-rich \emph{object sequence}. Concretely, we decompose the scene into object-level representations paired with object ID tokens, producing the scene embeddings consumed by the LLM (Figure~\ref{fig:model}). This section details these components. Based on this object-sequence design, we further introduce grounded CoT, which flexibly uses object IDs for multi-step reasoning over object relations (Section~\ref{sec:grounded_cot}).

% As shown in Figure~\ref{fig:model}, Chat-Scene++ decomposes the 3D scene into a set of object proposals, assigning each object a unique identifier. Multi-modal context-rich object features are extracted using large-scale pre-trained 3D and 2D models. These features are then mapped to the language model's embedding space, where scene embeddings are constructed by sequentially combining object identifiers with corresponding object-level embeddings. The resulting scene embeddings, along with system prompt embeddings and user prompt embeddings, are fed into the LLM to generate responses.

\noindent\textbf{Object identifier.}
To enable localized 3D scene understanding, we introduce a set of learnable identifier tokens $\{\texttt{<OBJ}i\texttt{>}\}_{i=1...n}$ to the LLM vocabulary. After tokenization, they yield object-ID embeddings $\{\mathbf{O}_i\}_{i=1...n}$.

% The identifier tokens are then integrated with the object tokens to establish one-to-one correspondences, enabling object referencing and grounding using these identifiers in subsequent conversations.

\begin{figure*}[t]
    \centering
    \includegraphics[width=0.95\textwidth]{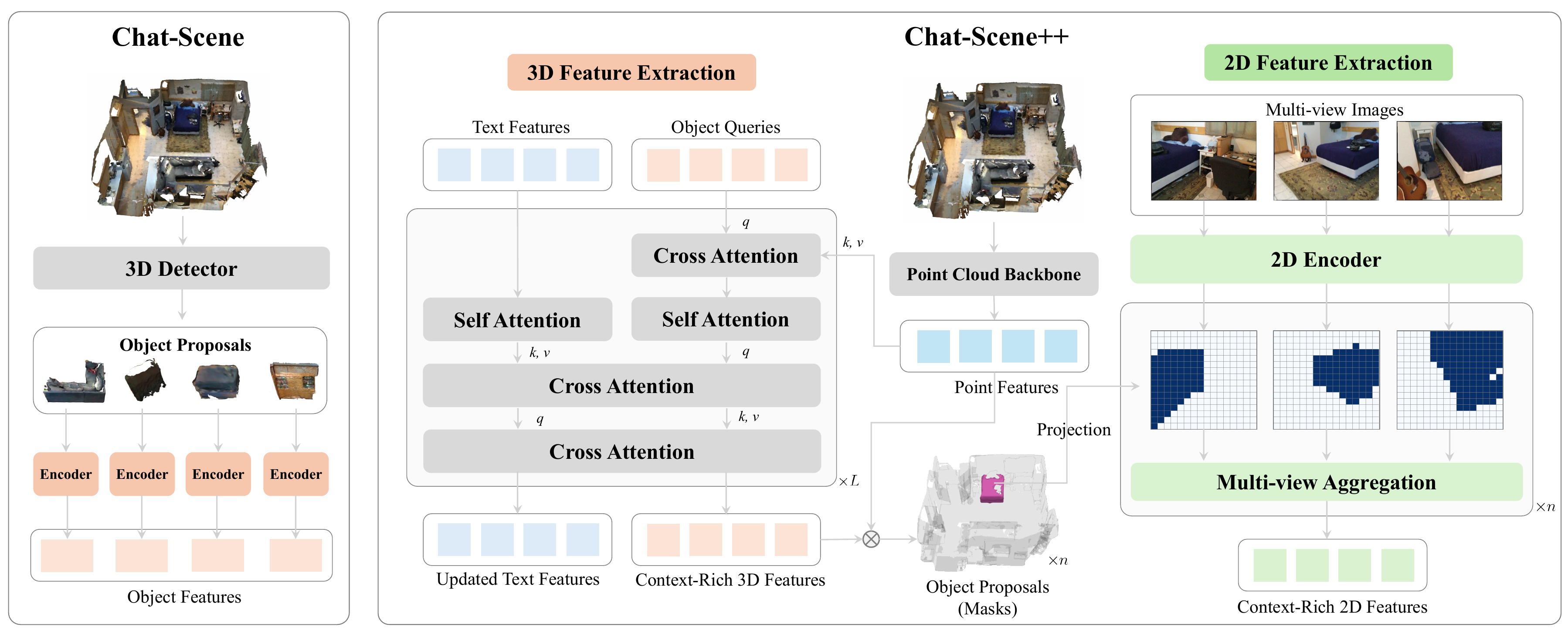}
    \caption{\textbf{Illustration of multi-modal context-rich feature extraction.} Chat-Scene~\cite{chatscene} extracts object-centric features for separate object proposals, while Chat-Scene++ extracts context-rich object features using 3D scene-level encoders and 2D image-level encoders.}
    \label{fig:feature_extract}
\end{figure*}

\noindent\textbf{Multi-modal context-rich object feature.}  
While object IDs provide a structured way to reference objects, extracting rich object features remains crucial. The scene-level representation requires a large amount of paired scene-language data for training, which is often unaffordable and labor-intensive due to the complexity of real-world scenarios. To address this challenge, our model represents the scene using a set of object-level embeddings, which obtain object representations from well-trained 2D and 3D models. In our previous work, Chat-Scene~\cite{chatscene}, we employed a two-stage pipeline: (i) using a 3D object detector Mask3D~\cite{mask3d} to generate object proposals, and (ii) using an object-level encoder Uni3D~\cite{uni3d} to extract object-centric features (Figure~\ref{fig:feature_extract}). However, this treats objects independently and misses contextual cues needed for scene understanding. To capture context in the object sequence, we adopt a multi-modal design that fuses 3D and 2D features using large-scale pre-trained 3D scene-level and 2D image-level encoders.

\noindent\textbf{\textit{3D context-rich feature.}}  
3D representations excel at capturing spatial relationships, geometric structures, and object shapes, which are essential for scene layout and depth understanding. To preserve inter-object context, one option is to use the hidden features of Mask3D~\cite{mask3d} object proposals directly; however, our ablations (Section~\ref{sec:ablate}) show this underperforms Uni3D~\cite{uni3d}. This is because Mask3D, trained for detection, mainly encodes category semantics, whereas Uni3D, pre-trained on large-scale 3D--text data, yields richer semantic representations.

Inspired by grounded pre-training~\cite{groundingdino, glip, grounded3dllm}, which aligns vision and text in transformer-based detectors to enrich object features, we follow CLASP (Contrastive Language-Scene Pre-training) from Grounded 3D-LLM~\cite{grounded3dllm} to pre-train a Mask3D~\cite{mask3d}-based architecture on a large-scale grounded scene--text dataset. This design jointly generates object proposals and extracts 3D object features using context-aware object queries from the scene-level transformer encoder.

As illustrated in Figure~\ref{fig:feature_extract}, the input set of $n$ object queries interact with text and point features through $L$ transformer layers, producing context-rich 3D features and masks for $n$ object proposals. For the $i$-th object, the context-rich feature is denoted as $\mathbf{Z}^{p}_{i}$, and its corresponding mask consists of $m_i$ points, where each point is represented by its 3D coordinate $(x, y, z)$, forming the point cloud $\mathbf{P}_i \in \mathbb{R}^{m_i \times 3}$.

\noindent\textbf{\textit{2D context-rich feature.}}  
2D representations provide rich appearance-based details, such as textures, colors, and fine-grained visual features, which are often more challenging to extract from raw 3D data. To enhance object representations, we extract context-rich 2D features from multi-view images. Specifically, we use the pre-trained 2D encoder DINOv2~\cite{dinov2} to extract local patch features. As shown in Figure~\ref{fig:feature_extract}, each image is first encoded to obtain hidden patch features with a patch size of $16 \times 16$. For the $i$-th object, we project its point cloud onto each image, transforming the projected mask to align with the patch size. \revcommentsecond{Added clarification of occlusion handing}\addedsecond{During projection, we leverage the depth annotations from ScanNet to determine visibility, so points that are occluded in a given view are not mapped to pixels and are naturally excluded from 2D feature extraction for that view.}
To aggregate patch features from multiple views, we proceed as follows:  

For each image, we compute a feature vector by averaging the features of masked patches, where the projected mask is nonzero:  
\begin{equation}
\mathbf{F}_i^{\text{img}} = \frac{1}{N_{\text{mask}}} \sum_{j \in \text{mask}} \mathbf{f}_j
\end{equation}  
where $\mathbf{f}_j$ denotes the feature of the $j$-th patch within the masked region, and $N_{\text{mask}}$ is the number of masked patches.  

We then aggregate features from all images, weighted by the mask size \( S_{\text{mask}} \) in each image, to obtain the final 2D feature of the $i$-th object:  
\begin{equation}
\mathbf{Z}_i^v = \frac{\sum_k \mathbf{F}_i^{\text{img}, k} \cdot S_{\text{mask}, k}}{\sum_k S_{\text{mask}, k}}
\end{equation}  
where $\mathbf{F}_i^{\text{img}, k}$ is the feature vector for the $k$-th image, and \( S_{\text{mask}, k} \) represents the mask size in the $k$-th image. Thus, the context-rich 2D feature of the $i$-th object is denoted as $\mathbf{Z}^v_i$.

\noindent\textbf{Feature-to-language projectors.}  
To align extracted object features with the language model, we introduce a 3D-language projector $f_p(\cdot)$ and a 2D-language projector $f_v(\cdot)$, which map 3D and 2D features into the language model's embedding space. For the $i$-th object, the projected token embeddings are given by:  
\begin{equation}
\mathbf{F}^{p}_{i} = f_p(\mathbf{Z}^{p}_{i}), \quad \mathbf{F}^{v}_{i} = f_v(\mathbf{Z}^{v}_{i}).
\end{equation}  
\noindent\textbf{Modeling scene as sequential context-rich objects.}  
Following the process described above, we obtain an object ID token embedding $\mathbf{O}_i$, a 3D object token embedding $\mathbf{F}^{p}_{i}$, and a 2D object token embedding $\mathbf{F}^{v}_{i}$ for each object. As shown in Figure~\ref{fig:model}, we construct the scene embeddings by sequentially combining each ID token embedding with its object token embeddings in a one-to-one manner. These embeddings are then fed into the LLM to represent the entire scene.

% \subsection{Prompt Template\label{sec:3d_llm_modeling}}
% Despite variations in task formulations, both referencing and grounding are unified using object identifiers. As illustrated in Table~\ref{tab:prompt_template}, the system message encodes object information in the scene as a sequence of ``$\texttt{<OBJ}i\texttt{>}$ $\texttt{<object>}$'', where $\texttt{<OBJ}i\texttt{>}$ denotes the identifier token for the $i$-th proposal, and $\texttt{<object>}$ serves as the placeholder for object tokens. The language tokenizer converts $\texttt{<OBJ}i\texttt{>}$ into its token embedding $\mathbf{O}_i$ and $\texttt{<object>}$ into the combined object token features $\mathbf{F}^{p}_{i}$ and $\mathbf{F}^{v}_{i}$. As illustrated by the following interaction, the user can directly employ identifier tokens to reference specific objects, while the assistant uses these tokens in responses to precisely ground target objects.

\subsection{3D LLM Modeling\label{sec:3d_llm_modeling}}  
Following prior multi-modal LLMs~\cite{3dllm, llava, leo}, we develop our 3D LLM using a visual instruction-tuning approach~\cite{llava}. Our model extends a pre-trained LLM~\cite{chiang2023vicuna} by integrating scene embeddings to enable multi-modal perception and 3D scene understanding. This section details our prompt template for interleaving scene information and interacting with the LLM via object IDs. We present a unified instruction-following framework for diverse tasks, facilitating joint training across various 3D scene understanding datasets.  

\begin{table}[htbp]
\centering
\caption{\textbf{Prompt template for the large language model.} }
\label{tab:prompt_template}
\begin{tcolorbox}[colback=lightgray!10,%gray background
                    colframe=black,% black frame colour
                    width=0.48\textwidth,
                    arc=1mm, auto outer arc,
                    boxrule=0.5pt,
                    ]
\textbf{System}: A chat between a curious user and an artificial intelligence assistant. The assistant gives helpful, detailed, and polite answers to the user's questions. The conversation centers around an indoor scene: [$\texttt{<OBJ001>}$ $\texttt{<object>}$ $\texttt{<OBJ002>}$ $\texttt{<object>}$ ... $\texttt{<OBJ}n\texttt{>}$ $\texttt{<object>}$]. 

\textbf{User:} Find the closest trash bin to $\texttt{<OBJ013>}$.

\textbf{Assistant:} There are two trash bins, $\texttt{<OBJ023>}$ and $\texttt{<OBJ032>}$, both located near the chair.

\end{tcolorbox}
\label{tab:prompt_input}
\end{table}

% \noindent\textbf{Prompt template.}  
% As shown in Table~\ref{tab:prompt_template}, the system message encodes scene information as a sequence of ``$\texttt{<OBJ}i\texttt{>}$ $\texttt{<object>}$,'' where $\texttt{<OBJ}i\texttt{>}$ represents the ID token for the $i$-th object proposal, and $\texttt{<object>}$ acts as a placeholder for object tokens. The language tokenizer maps $\texttt{<OBJ}i\texttt{>}$ to its token embedding $\mathbf{O}_i$ and $\texttt{<object>}$ to the combined object token features $\mathbf{F}^{p}_{i}$ and $\mathbf{F}^{v}_{i}$. In the following conversation example, the user can directly reference objects using ID tokens, while the assistant leverages these tokens in responses to ensure precise object grounding.  
\noindent\textbf{Prompt template.}  
As shown in Table~\ref{tab:prompt_template}, the system message encodes scene information as a sequence of ``$\texttt{<OBJ}i\texttt{>}$ $\texttt{<object>}$,'' where $\texttt{<OBJ}i\texttt{>}$ represents the ID token for the $i$-th object proposal, and $\texttt{<object>}$ acts as a placeholder for object tokens. \revcommentsecond{Added clarification of ID order}\addedsecond{The specific numbering of ID tokens is arbitrary; the model does not learn semantics from the ID order itself. To encourage order-invariance, we intentionally randomize the object order during training, so objects can be numbered in any sequence at inference with negligible impact.} The language tokenizer maps $\texttt{<OBJ}i\texttt{>}$ to its token embedding $\mathbf{O}_i$ and $\texttt{<object>}$ to the combined object token features $\mathbf{F}^{p}_{i}$ and $\mathbf{F}^{v}_{i}$. In the following conversation example, the user can directly reference objects using ID tokens, while the assistant leverages these tokens in responses to ensure precise object grounding.

\begin{figure*}[tbp]
    \centering
    \includegraphics[width=0.8\textwidth]{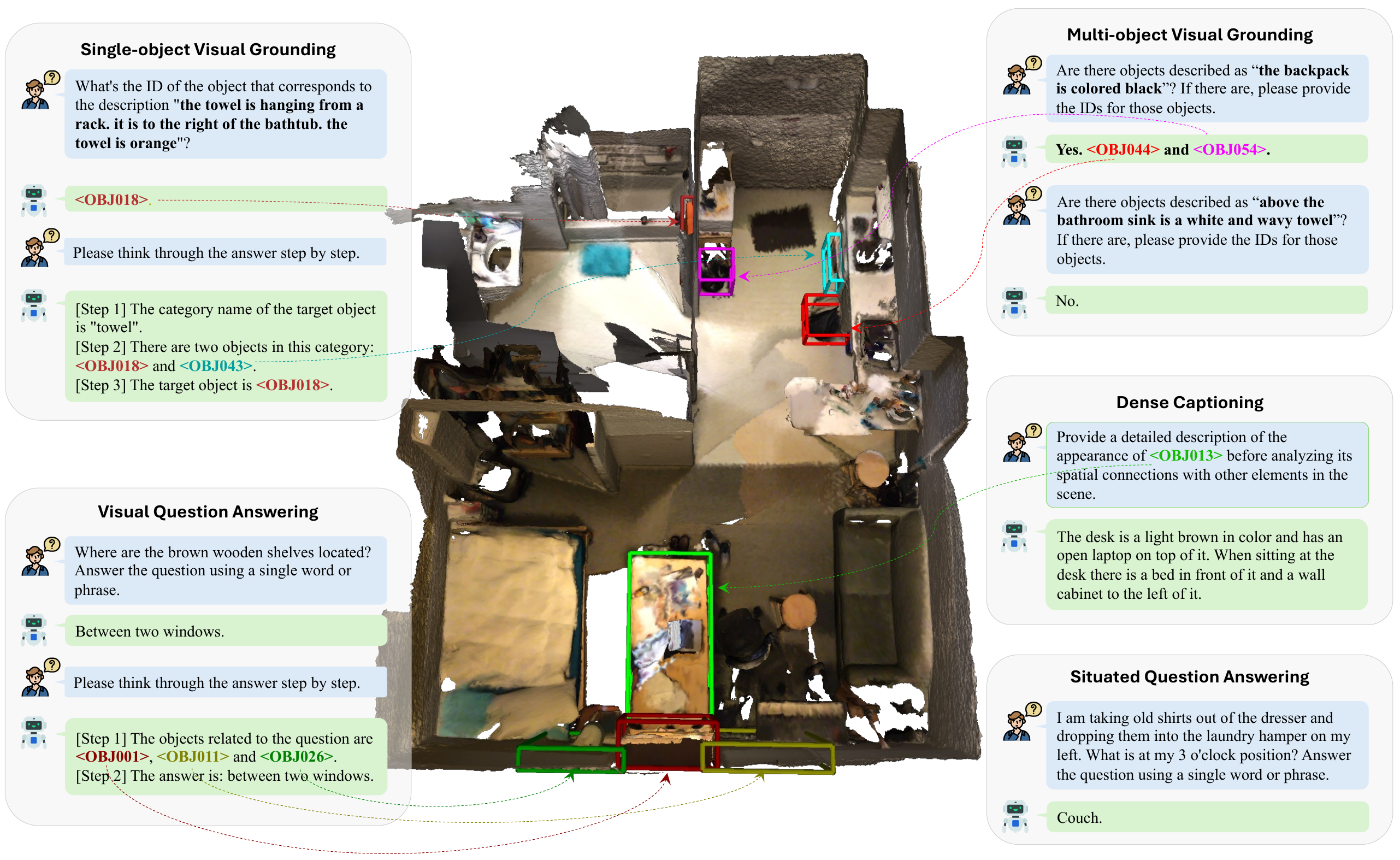}
    \caption{\textbf{Examples of various 3D scene-language understanding tasks}. All the tasks are unified to single-turn question-answering pairs without extra task heads. Object identifiers are used to reference and ground the object during the conversation.}
    \label{fig:taskflow}
\end{figure*}

\noindent\textbf{Unified instruction-following for diverse tasks.}  
Unlike previous 3D LLMs that require a dedicated prompt encoder for user-specified objects~\cite{ll3da} or an additional head for object grounding~\cite{grounded3dllm}, we unify all downstream tasks into a consistent instruction-following framework. These tasks include 3D visual grounding (ScanRefer \& Multi3DRef), 3D dense captioning (Scan2Cap), and 3D visual question answering (ScanQA \& SQA3D). Each data sample follows a single-turn user-assistant interaction format, as illustrated in Figure~\ref{fig:taskflow}. This unified approach enables joint training across all tasks without requiring task-specific heads. Moreover, it allows seamless augmentation of existing datasets with additional 3D scene-language data, enhancing scene-language alignment, as summarized in Table~\ref{tab:data_statistics}.

\subsection{Grounded Chain-of-Thought (G-CoT)}\label{sec:grounded_cot}  
The sequential object representations in Chat-Scene++ enable flexible multi-step inference, allowing the model to reason about object relationships in 3D scenes. However, in current 3D vision-language tasks such as ScanQA~\cite{scanqa} and ScanRefer~\cite{scanrefer}, target answers are often overly concise, providing limited supervision for training LLMs to perform complex reasoning. In ScanQA, answers are typically single words or phrases, while in ScanRefer, they are single object IDs (within our framework). This form of supervision lacks the granularity needed for effective reasoning. For example, if the model is asked to count the number of chairs in a scene and answers ``4'', it does not specify which four objects were identified as chairs, leaving the reasoning process entirely absent.

To address this limitation, we propose \textbf{grounded chain-of-thought} (\textit{G-CoT}), which explicitly incorporates relevant object information into the reasoning process. While recent methods~\cite{llava-cot, insight-v} have explored CoT for multi-modal LLMs, they primarily focus on text-based reasoning. In contrast, our approach leverages object IDs to anchor reasoning steps to specific objects within the scene, enabling more interpretable and grounded multi-step inference. 

We design grounded CoT in a structured, multi-step format tailored to the specific requirements of each dataset. For ScanQA, which provides annotations for related objects, the reasoning process consists of two steps: (1) predicting the objects relevant to the question and (2) inferring the final answer based on these objects. This semantic-level reasoning leverages ScanQA’s annotations to ensure that the model focuses on semantically related objects, making it easier to analyze and derive the correct answer.

For ScanRefer, where related object annotations are unavailable, we explore two forms of reasoning to enhance object localization: \textit{category-level} and \textit{space-level}. For category-level reasoning, we consider all objects in the same category as the target object. The reasoning process follows three steps: (1) predicting the target object’s category (a common task in expert models~\cite{scanrefer, 3djcg, 3d-vlp}), (2) identifying all objects in this category, and (3) locating the target object. This approach helps the model narrow down the search space by focusing on semantically similar objects, which is particularly useful when the scene contains multiple instances of the same object type. For space-level reasoning, we define the top-5 closest objects to the target as related. The reasoning process follows two steps: (1) identifying the top-5 closest objects and (2) locating the target object. This approach leverages spatial proximity to capture local context, which can be effective in cluttered scenes. By exploring these two forms of reasoning, we aim to understand their respective strengths and limitations in improving the model’s ability to reason about object relationships and accurately locate the target.

Using the original annotations from ScanQA and ScanRefer, we automatically generate grounded CoT data through predefined templates, as shown in Table~\ref{tab:prompt_template_gcot}. Experiments on both datasets demonstrate the effectiveness of grounded CoT for VQA and visual grounding tasks. For captioning tasks in Scan2Cap~\cite{scan2cap}, where the target object is explicitly specified in the prompt, grounded CoT is unnecessary for object identification, as the focus shifts to generating descriptive text rather than reasoning about object relationships.

\revcommentsecond{Added discussion}\addedsecond{We note that the category-level and space-level templates are introduced only for constructing supervised G-CoT training data, where they provide a simple and controllable way to generate grounded rationales and isolate the effect of grounded supervision. During \emph{inference}, our method does not select among multiple hand-crafted templates; instead, we use a single generic prompt (``please think through the answer step by step'') to elicit grounded chain-of-thought reasoning, following standard CoT prompting practice.}

\begin{table}[tbp]
\centering
\caption{\textbf{Templates for grounded chain-of-thought.} }
\label{tab:prompt_template_gcot}
\begin{tcolorbox}[colback=lightgray!10,%gray background
                    colframe=black,% black frame colour
                    width=0.48\textwidth,
                    arc=1mm, auto outer arc,
                    boxrule=0.5pt,
                    ]
{\small \textbf{\textit{Grounded CoT for VQA}}}

\textbf{User:} <\textit{Question}>. Please think through the answer step by step.

\textbf{Assistant:} 

[Step 1] The objects related to the question is/are \texttt{<OBJXXX>}, ... \texttt{<OBJXXX>}, and \texttt{<OBJXXX>}.

[Step 2] The answer is: <\textit{Answer}>.
\\
\\
{\small \textbf{\textit{Grounded CoT for Visual Grounding}}}

\textbf{User:} What's the ID of the object that corresponds to the description ``<\textit{Description}>''? Please think through the answer step by step.

\textbf{Assistant:} \textcolor{blue}{\textit{(Category-level)}}

[Step 1] The category name of the target object is ``<\textit{Category}>''.

[Step 2] There are $M$ objects in this category: \texttt{<OBJXXX>}, ... \texttt{<OBJXXX>}, and \texttt{<OBJXXX>}.

[Step 3] The target object is \texttt{<OBJXXX>}.

\textbf{Assistant:} \textcolor{blue}{\textit{(Space-level)}}

[Step 1] The spatially related objects could include: \texttt{<OBJXXX>}, ... \texttt{<OBJXXX>}, and \texttt{<OBJXXX>}.

[Step 2] The target object is \texttt{<OBJXXX>}.

\end{tcolorbox}
\end{table}

\subsection{Training Strategy\label{sec:training_strategy}}

\noindent\textbf{Training objective\label{sec:training_and_inference}}.
We have unified all tasks into a consistent instruction-following format. As a result, the only training loss during the joint-training phase is the Cross-Entropy loss of the language model. The training objective is to optimize the trainable parameters, denoted by $\theta$, with the goal of minimizing the negative log-likelihood of the target response sequence $s^\mathrm{res}$ generated by the assistant. Specifically, given the input prefix sequence $s^\mathrm{prefix}$, which includes both system messages and user instructions, the loss function is defined as follows:
\begin{equation}
\mathcal{L}(\theta) = -\sum_{i=1}^{k} \log P(s^\mathrm{res}_{i} | s^\mathrm{res}_{[1, \ldots, i-1]}, s^\mathrm{prefix}),
\end{equation}
where $k$ is the number of tokens in the response sequence, and $s^\mathrm{res}_{[1, \ldots, i-1]}$ represents the sequence of the previous $i-1$ tokens in the response. The set of trainable parameters $\theta$ includes two vision-language projectors, $n$ newly added token embeddings for object IDs, and the language model itself.

\noindent\textbf{One-stage joint training.}
Most existing MLLMs~\cite{scenellm, llava, leo} adopt a two-stage training approach, consisting of an initial alignment phase to train the projector exclusively, followed by a fine-tuning phase for both the projector and the language model. This approach not only requires additional data and extended training time for alignment but also introduces complexity in determining the optimal duration for the initial phase. In contrast, we opt for a one-stage training process, where both the projectors and the language model are trained simultaneously. Our experiments show that this jointly trained model performs exceptionally well, eliminating the need for separate fine-tuning on specific downstream tasks.

\revcommentfirst{We add more training details in Appendix A.}

\revcommentsecond{We add more details of training and inference costs in Appendix A.}

\section{Experiments\label{sec:experiment}}

\begin{table}[tbp]
\centering
\caption{\textbf{Statistics of training data.}} \label{tab:data_statistics}
\resizebox{0.8\linewidth}{!}{
\begin{tabular}{ccc}
\toprule
Training Data & Source & \# Sample \\ \midrule
ScanRefer & ScanRefer~\cite{scanrefer} & 33K \\
ScanRefer\_CoT & ScanRefer~\cite{scanrefer} & 24K \\
Multi3DRefer & Multi3DRefer~\cite{multi3drefer} & 39K \\
Scan2Cap & Scan2Cap~\cite{scan2cap} & 33K \\
ScanQA & ScanQA~\cite{scanqa} & 26K \\
ScanQA\_CoT & ScanQA~\cite{scanqa} & 23K \\
SQA3D & SQA3D~\cite{sqa3d} & 26K \\
ObjAlign & ScanNet~\cite{scannet} & 25K \\
Nr3DCaption & Nr3D~\cite{referit3d} & 28K \\
ObjCaption & Grounded 3D-LLM~\cite{grounded3dllm} & 24K \\
SceneCaption & Grounded 3D-LLM~\cite{grounded3dllm} & 43K \\
\textbf{Total} & \textbf{--} & \textbf{329K} \\ \bottomrule
\end{tabular}}
\end{table}

\subsection{Datasets and Metrics\label{sec:datasets}}

\noindent \textbf{Datasets.}
We conducted experiments on five benchmarks: ScanRefer~\cite{scanrefer} for single-object visual grounding, Multi3DRefer~\cite{multi3drefer} for multi-object visual grounding, Scan2Cap~\cite{scan2cap} for dense captioning, and ScanQA~\cite{scanqa} and SQA3D~\cite{sqa3d} for visual question answering. These benchmarks are based on the ScanNet dataset~\cite{scannet}, which contains 1,513 richly annotated RGB-D scans of real-world indoor scenes. All benchmarks follow the same train/validation/test splits, enabling joint training and evaluation.

To enhance alignment, we augment the original training sets with additional datasets, resulting in a total of 329K training samples, as summarized in Table~\ref{tab:data_statistics}. ObjAlign data is automatically generated using ScanNet's class annotations, requiring the model to predict object class labels. Nr3DCaption is an object-captioning dataset derived from the caption annotations in Nr3D~\cite{referit3d}. SceneCaption and ObjCaption, sourced from Grounded 3D-LLM~\cite{grounded3dllm}, provide detailed scene- and object-level descriptions.

\noindent \textbf{Metrics.}
We follow the standard evaluation metrics used in these benchmarks. For ScanRefer~\cite{scanrefer}, we measure thresholded accuracy using Acc@0.25 and Acc@0.5, where predictions are considered correct if their Intersection over Union (IoU) with ground truth exceeds 0.25 or 0.5, respectively. Multi3DRefer~\cite{multi3drefer} evaluates multi-object grounding using the F1 score at IoU thresholds of 0.25 and 0.5. For Scan2Cap~\cite{scan2cap}, we use CIDEr@0.5 and BLEU-4@0.5 (C@0.5 and B-4@0.5), combining image captioning metrics with IoU-based spatial alignment. ScanQA~\cite{scanqa} is assessed with CIDEr~\cite{cider} and BLEU-4~\cite{bleu}, abbreviated as C and B-4. SQA3D~\cite{sqa3d} is evaluated using exact match accuracy (EM) and its refined version, EM-R, as proposed by LEO~\cite{leo}.

\begin{table*}[t]
\centering
\caption{\textbf{Performance comparison.} ``Expert models'' are tailored for specific tasks using task-oriented heads, while ``LLM-based models'' are designed for general instructions and responses. Entries in {\color{lightgray}gray} denote the incorporation of ground-truth objects into the model inputs.} \label{tab:performance_comparison}
\resizebox{\linewidth}{!}{
\begin{tabular}{c|ccc|cccccccccc}
\toprule
 &
  \multirow{2}{*}{Method} &
  \multirow{2}{*}{2D} & 
  \multirow{2}{*}{LLM} & 
  \multicolumn{2}{c}{ScanRefer} &
  \multicolumn{2}{c}{Multi3DRefer} &
  \multicolumn{2}{c}{Scan2Cap} &
  \multicolumn{2}{c}{ScanQA} &
  \multicolumn{2}{c}{SQA3D} \\ 
 &                 & & & Acc@0.25 & Acc@0.5 & F1@0.25 & F1@0.5 & C@0.5 & B-4@0.5 & C & B-4 & EM   & EM-R \\ \midrule
\multirow{9}{*}{\rotatebox[origin=c]{90}{Expert Models}}   & ScanRefer~\cite{scanrefer}        & \xmark    & --    & 37.3     & 24.3    & --       & --      & --         & --          & --     & --      & --    & --          \\
 & ViL3DRel~\cite{vil3drel}           & \xmark   & --    & 47.9     & 37.7    & --       & --      & --      & --       & --  & --   & --    & --          \\
 & ScanQA~\cite{scanqa}           & \cmark   & --    & --        & --       & --       & --      & --         & --          & 64.9  & 10.1   & --    & --          \\
 & 3DJCG~\cite{3djcg}            & \cmark   & --    & 49.6     & 37.3    & --       & --      & 49.5      & 31.0       & --     & --      & --    & --          \\
 & 3D-VLP~\cite{3d-vlp}           & \cmark   & --    & 51.4     & 39.5    & --       & --      & 54.9      & 32.3       & 67.0  & 11.1   & --    & --          \\
 & M3DRef-CLIP~\cite{multi3drefer}      & \cmark   & --    & 51.9     & 44.7    & 42.8    & 38.4   & --         & --          & --     & --      & --    & --          \\
 & 3D-VisTA~\cite{3dvista}         & \xmark   & --    & 50.6     & 45.5    & --       & --      & 66.9      & 34.0       & 72.9  & 13.1   & 48.5 & --          \\
 & Vote2Cap-DETR++~\cite{vote2cap++}  & \cmark   & --    & --        & --       & --       & --      & 67.6      & 37.1       & --     & --      & --    & --          \\
 & ConcreteNet~\cite{concretenet}      & \xmark   & --    & 50.6     & 46.5    & --       & --      & --         & --          & --     & --      & --    & --          \\
 \hline
 \\ [-2ex]
 \multirow{9}{*}{\rotatebox[origin=c]{90}{LLM-based Models}} & Chat-3D~\cite{chat3d}           & \xmark   & Vicuna-7B   & --        & --       & --       & --      & --         & --          & 53.2  & 6.4    & --    & --          \\
 & 3D-LLM~\cite{3dllm}           & \cmark   & Flan-T5-XL 3B   & 30.3     & --       & --       & --      & --         & --          & 69.4  & 12.0   & --    & --          \\
 & LL3DA~\cite{ll3da}            & \xmark   & OPT-1.3B   & --        & --       & --       & --      & 65.2      & 36.8       & 76.8  & 13.5   & --    & --          \\
 & LEO~\cite{leo}               & \xmark   & Vicuna-7B   & --        & --       & --       & --      & 72.4      & 38.4       & {\color{lightgray}101.4}  & {\color{lightgray}13.2}   & --    & 53.7       \\
 & Scene-LLM~\cite{scenellm}        & \cmark   & Llama2-7B   & --        & --       & --       & --      & --         & --          & 80.0  & 12.0   & 54.2 & --          \\
 & Grounded 3D-LLM~\cite{grounded3dllm} & \xmark   & Vicuna-7B   & 48.6     & 44.0    & 44.7    & 40.6   & 70.2      & 35.0       & 75.9  & 13.2   & --    & --          \\
 & Chat-Scene~\cite{chatscene}       & \cmark   & Vicuna-7B   & 55.5     & 50.2    & 57.1    & 52.4   & 77.1      & 36.3       & 87.7  & 14.3   & 54.6 & 57.5     \\
 & Chat-Scene++       & \xmark   & Vicuna-7B & 53.0     & 47.3    & 56.4    & 51.6   & 72.8      & 34.6       & 82.5  & 13.1   & 54.2 & 57.3   \\
 & Chat-Scene++       & \cmark   & Vicuna-7B & \textbf{59.8}     & \textbf{53.4}    & \textbf{62.9}    & \textbf{58.7}   & \textbf{83.0}      & \textbf{38.7}       & \textbf{93.5}  & \textbf{18.4}   & \textbf{55.8} & \textbf{58.5}   \\
 \bottomrule 
\end{tabular}}

\end{table*}

\subsection{Comparison with State-of-the-Art Methods}

\noindent \textbf{Compared baselines.}  
The baseline models fall into two main categories: traditional expert models and general multi-modal large language models. Expert models generate structured outputs tailored to specific tasks, while LLM-based models produce open-ended responses applicable to a broader range of scenarios.

\begin{itemize}[leftmargin=*]
    \item \textbf{Expert models:}  
    Traditional expert models serve as strong baselines for specific benchmarks. ScanRefer~\cite{scanrefer} and ScanQA~\cite{scanqa} establish foundational performance on their respective datasets. 3DJCG~\cite{3djcg} integrates multiple tasks into a unified architecture, leveraging the synergy between visual grounding and dense captioning. Both 3D-VLP~\cite{3d-vlp} and 3D-VisTA~\cite{3dvista} develop pre-training strategies for aligning 3D scenes with language, enhancing multi-task capabilities. M3DRef-CLIP~\cite{multi3drefer} advances multi-object grounding while also improving single-object grounding performance. ConcreteNet~\cite{concretenet}, a state-of-the-art model for ScanRefer, introduces novel verbo-visual fusion techniques to enhance dense 3D visual grounding. Vote2Cap-DETR++~\cite{vote2cap++}, the leading model on Scan2Cap, employs parallel decoding to decouple caption generation from object localization.

    \item \textbf{LLM-based models:}  
    Chat-3D~\cite{chat3d} adopts an object-centric approach but lacks generalizability to broader 3D scene tasks. 3D-LLM~\cite{3dllm} incorporates location tokens for object grounding but is constrained by data limitations. LL3DA~\cite{ll3da} introduces an assistant that directly processes point cloud data, responding to textual and visual prompts. LEO~\cite{leo} pioneers an embodied generalist approach by integrating action tokens, while Scene-LLM~\cite{scenellm} enhances 3D reasoning with a combination of scene-level and egocentric information. However, LL3DA, LEO, and Scene-LLM lack explicit grounding capabilities. Grounded 3D-LLM~\cite{grounded3dllm} addresses this by learning referent tokens for object proposals but requires an additional contrastive loss for alignment.
\end{itemize}

\noindent \textbf{Analysis.} 
As shown in Table~\ref{tab:performance_comparison}, our model outperforms previous methods across all metrics without task-specific fine-tuning, demonstrating a promising unified framework for 3D scene understanding. In visual grounding tasks, our model improves the state-of-the-art (SOTA) performance by 3.2\% (Acc@0.5) on ScanRefer and 7.1\% (F1@0.5) on Multi3DRefer, showcasing its strong grounding capabilities. For dense captioning, it achieves a 5.9\% improvement in SOTA performance (CIDEr@0.5) on Scan2Cap, highlighting its effectiveness in object referring and captioning. In VQA tasks on ScanQA and SQA3D, which do not require object referencing or grounding, our model enhances SOTA performance by 5.8\% (CIDEr) on ScanQA and 1.2\% (EM) on SQA3D, demonstrating superior overall 3D scene understanding and reasoning.

\revcommentfirst{We have expanded our failure case analysis in Appendix D. The failures are now categorized into several distinct types to provide deeper insights into the model's limitations and to guide future improvements.}

\subsection{Ablation Study}
\label{sec:ablate}
\noindent \textbf{Effects of object identifiers.}  
Table~\ref{tab:identifier_ablate} demonstrates that the format of object identifiers (IDs) affects both performance and token cost. When comparing token costs, we consider all tokens used to represent objects, including object IDs and features. The number of objects is denoted as $N$.  

We first evaluate a setting without object IDs, where visual grounding follows 3D-LLM’s method~\cite{3dllm} using location tokens to predict bounding boxes. This results in a significant drop in visual grounding performance on ScanRefer and Multi3DRefer, highlighting the importance of object IDs.

Next, we evaluate the impact of different ID formats. A straightforward approach is to use plain text IDs such as ``\texttt{Obj001}'', which is tokenized into four tokens (``\texttt{Obj}'', ``\texttt{0}'', ``\texttt{0}'', ``\texttt{1}''). Including two feature object tokens (3D \& 2D), representing a single object requires six tokens in total. This high token cost increases training and inference time, making the approach impractical for real-world applications. To mitigate this, we explore using a single token per ID by adding new tokens to the language model’s vocabulary. We assess two strategies: employing fixed random Gaussian embeddings (``Gaussian'') and using learnable tokens (``Learnable''). The results show that learnable tokens enhance performance while reducing token costs. Lowering token costs from $6N$ to $3N$ significantly improves efficiency, particularly when handling 3D scenes with hundreds of objects.  

Lastly, we assess the effect of ID ordering during training. Randomizing ID order acts as data augmentation, improving overall performance by mitigating imbalances in object ID distributions, particularly in ground-truth annotations for visual grounding tasks.  

\begin{table*}[tbp]
\centering
\caption{\textbf{Ablation studies on object identifiers.} ``Plain Text'' employs plain text for object numbers, ``Gaussian'' uses fixed Gaussian embeddings, and ``Learnable'' learns new identifier tokens. ``Random Order'' means using random ID order during training. ``Token Cost'' denotes the total tokens for $N$ objects, including object identifiers.} \label{tab:identifier_ablate}
\resizebox{0.85\linewidth}{!}{
\begin{tabular}{cccc|ccccc}
\toprule
\multirow{2}{*}{\#} & \multirow{2}{*}{ID Type}  & \multirow{2}{*}{ID Order} & \multirow{2}{*}{\makecell{Token\\Cost}} & ScanRefer & Multi3DR. & Scan2Cap & ScanQA & SQA3D \\
& &  &  & Acc@0.5 & F1@0.5 & C@0.5 & CIDEr & EM \\ 
 \midrule
1 & \xmark & \xmark & 2$N$ & 25.5 & 29.9 & 52.8 & 78.2 & 50.0 \\
2 & Plain Text & Fixed & 6$N$ & 49.2 & 54.8 & 77.9 & 89.0 & 55.1 \\ 
3 & Gaussian  & Fixed & 3$N$ & 48.3 & 54.5 & 76.4 & 88.1 & 54.9 \\
4 & Learnable & Fixed & 3$N$ & 52.6 & 57.7 & 81.1 & 93.0 & 55.7 \\
5 & Learnable & Random & 3$N$ & \textbf{53.4} & \textbf{58.7} & \textbf{83.0} & \textbf{93.5} & \textbf{55.8} \\
\bottomrule
\end{tabular}}
\end{table*}

\begin{table*}[tbp]
\centering
% \caption{\textbf{Ablation studies on feature extraction.} ``object-level'' represents the use of object-level encoders. ``Pos.'' adds additional position embedding to introduce the spatial information. ``Scene-level'' uses scene-level encoders.} \label{tab:encoder_ablate}
\caption{\textbf{Ablation studies on object features.} ``Pos.'' adds additional position embeddings to introduce spatial context.} \label{tab:encoder_ablate}
\resizebox{0.85\linewidth}{!}{
\begin{tabular}{ccc|ccccc}
\toprule
\multirow{2}{*}{\#} & \multirow{2}{*}{3D} & \multirow{2}{*}{2D} & ScanRefer & Multi3DR. & Scan2Cap & ScanQA & SQA3D \\
 &  &  & Acc@0.5 & F1@0.5 & C@0.5 & CIDEr & EM \\ \midrule
1 & Mask3D~\cite{mask3d} & \xmark & 37.6 & 40.9 & 60.5 & 76.2 & 52.2 \\
2 & Uni3D~\cite{uni3d} & \xmark & 41.2 & 43.8 & 64.9 & 80.3 & 53.4 \\
3 & Uni3D~\cite{uni3d} + Pos. & \xmark & 41.5 & 44.1 & 64.8 & 79.9 & 53.3 \\
4 & Uni3D~\cite{uni3d} + Pos. & DINOv2~\cite{dinov2} & 50.7 & 53.2 & 78.5 & 88.2 & 54.7 \\
5 & Mask3D w/ CLASP~\cite{grounded3dllm} & \xmark & 47.3 & 51.6 & 72.8 & 82.5 & 54.2 \\
6 & Mask3D w/ CLASP~\cite{grounded3dllm} & CLIP~\cite{CLIP} & 49.2 & 54.5 & 78.6 & 89.9 & 54.1 \\ 
7 & Mask3D w/ CLASP~\cite{grounded3dllm} & DINOv2~\cite{dinov2} & \textbf{53.4} & \textbf{58.7} & \textbf{83.0} & \textbf{93.5} & \textbf{55.8} \\ 
\bottomrule
\end{tabular}}
\end{table*}

% \noindent \textbf{Object Identifiers.}  
% Table~\ref{tab:identifier_ablate} illustrates how the format of object identifiers impacts both performance and token cost. Token cost accounts for all tokens representing objects, including object IDs and features, with the number of objects denoted as $N$.  

% We first evaluate a setting without object identifiers, where visual grounding follows 3D-LLM’s method~\cite{3dllm} using location tokens to predict bounding boxes. This results in a significant drop in visual grounding performance on ScanRefer and Multi3DRefer, highlighting the importance of object identifiers.  

% Next, we examine different identifier formats. A naive approach uses plain text identifiers (e.g., ``Obj001''), which are tokenized into four tokens (``Obj'', ``0'', ``0'', ``1''). With two feature tokens (3D \& 2D), each object requires six tokens, leading to high computational costs. To reduce this, we explore single-token identifiers by expanding the language model’s vocabulary. We compare two strategies: fixed random Gaussian embeddings (``Gaussian'') and learnable tokens (``Learnable''). Results show that learnable tokens improve performance while cutting token costs from $6N$ to $3N$, reducing memory usage and accelerating training and inference—especially for large 3D scenes.  

% Lastly, we assess the effect of identifier ordering during training. Randomizing identifier order acts as data augmentation, improving overall performance by mitigating imbalances in object ID distributions, particularly in ground-truth annotations for visual grounding tasks.  

\begin{table}[tbp]
\centering
\caption{\textbf{Ablation studies on grounded CoT.} ``QA CoT'' uses CoT data for visual question answering task ScanQA. ``Grounded CoT'' uses CoT data for visual grounding task ScanRefer. ``Two-stage'' involves an additional training stage using CoT data. ``Space-level'' and ``Category-level'' denote two different templates for grounded CoT.} \label{tab:cot_ablate}
\resizebox{\linewidth}{!}{
\begin{tabular}{cccc|cccc}
\toprule
\multirow{2}{*}{\#} & \multirow{2}{*}{\makecell{QA\\CoT}} & \multirow{2}{*}{\makecell{Grounded\\CoT}} & \multirow{2}{*}{\makecell{Training\\Schema}} & \multicolumn{2}{c}{ScanQA} & \multicolumn{2}{c}{ScanRefer} \\
 &  &  &  & CIDEr & B-4 & Acc@0.25 & Acc@0.5 \\ \midrule
1 & \xmark & \xmark & One-stage & 89.3 & 15.6 & 57.2 & 52.1 \\
2 & \cmark & \xmark & One-stage & 93.1 & 17.9 & 57.3 & 52.0 \\
3 & \xmark & Space-level & One-stage & 91.2 & 15.4 & 58.0 & 51.8 \\
4 & \xmark & Category-level & One-stage & 90.2 & 15.9 & 59.3 & 52.9 \\
5 & \cmark & Category-level & One-stage & \textbf{93.5} & \textbf{18.4} & 59.8 & 53.4 \\
6 & \cmark & Category-level & Two-stage & 92.9 & 17.6 & \textbf{60.3} & \textbf{53.8} \\ \bottomrule
\end{tabular}}
\end{table}

\begin{table*}[tbp]
\centering
\caption{\textbf{Ablation studies on LLM backbones.}} \label{tab:llm_ablate}
\resizebox{0.75\linewidth}{!}{
\begin{tabular}{c|ccccc}
\toprule
\multirow{2}{*}{LLM} & ScanRefer & Multi3dRefer & Scan2Cap & ScanQA & SQA3D \\
 & Acc@0.5 & F1@0.5 & C@0.5 & CIDEr & EM \\ \midrule
OPT-1.3B~\cite{opt1.3b} & 50.5 & 54.1 & 81.3 & 88.7 & 54.5 \\
Tiny-Vicuna-1B~\cite{chiang2023vicuna} & 52.6 & 56.1 & 81.9 & 91.7 & 54.7 \\
Vicuna-7B~\cite{chiang2023vicuna} & 53.4 & 58.7 & 83.0 & 93.5 & 55.8 \\
Vicuna-13B~\cite{chiang2023vicuna} & 53.0 & 57.9 & 84.2 & \textbf{94.2} & 56.1 \\
LLaMA3-8B~\cite{llama3} & \textbf{55.7} & \textbf{60.4} & \textbf{84.7} & 91.0 & \textbf{56.8} \\ \bottomrule
\end{tabular}}
\end{table*}

\begin{table}[tp]
\caption{\revcommentsecond{Revised table}\addedsecond{\textbf{Zero-shot grounded CoT results on SQA3D.} We report per-type accuracy and the paired-bootstrap 95\% CI of $\Delta$ (w/ G-CoT $-$ w/o G-CoT).}}
\label{tab:sqa3d}
\resizebox{0.98\linewidth}{!}{
\begin{tabular}{lccccccc}
\toprule
Method & What & Is & How & Can & Which & Others & Avg. \\
\midrule
w/o G-CoT & 46.9 & 67.9 & \textbf{53.8} & \textbf{71.2} & 50.0 & 56.2 & 55.8 \\
w/ G-CoT  & \textbf{47.1} & \textbf{69.2} & 53.3 & 71.0 & \textbf{52.6} & \textbf{56.7} & \textbf{56.4} \\
$\Delta$ (95\% CI) &
+0.2$\pm$0.1 &
+1.3$\pm$0.3 &
--0.5$\pm$0.2 &
--0.2$\pm$0.1 &
+2.5$\pm$0.5 &
+0.5$\pm$0.2 &
+0.6$\pm$0.1 \\

\bottomrule
\end{tabular}}
\end{table}

\begin{table}[ht]
\centering
\caption{\revcommentfirst{New table}\addedfirst{\textbf{Open-vocabulary results on 3RScan~\cite{3rscan}}}}
\label{tab:3rscan_results}
\begin{tabular}{cccc}
\toprule
Method & 3RQA & 3RDialog & 3RPlan \\
\midrule
LEO (zero-shot) & 35.8 & 25.5 & 23.4 \\
\textbf{Ours (zero-shot)} & \textbf{38.2} & \textbf{34.3} & \textbf{35.0} \\
LEO (fine-tuned) & 51.9 & 73.3 & 81.1 \\
\textbf{Ours (fine-tuned)} & \textbf{58.1} & \textbf{85.7} & \textbf{96.5} \\
\bottomrule
\end{tabular}
\end{table}

\begin{table}[t]
\centering
\caption{\revcommentsecond{New Table}\addedsecond{Zero-shot evaluation on the NuScenes-QA~\cite{nuscenes-qa} validation set. Methods are grouped by training setting (finetuned vs.\ zero-shot). Best zero-shot results are bolded.}}
\label{tab:nuscene-qa}
\resizebox{\linewidth}{!}{
\begin{tabular}{l cc ccc c}
\toprule
\textbf{Method} & \textbf{Exist} & \textbf{Count} & \textbf{Object} & \textbf{Status} & \textbf{Comparison} & \textbf{Accuracy} \\
\midrule
\multicolumn{7}{l}{\textbf{\textit{Finetuned}}} \\
LiDAR-LLM~\cite{lidar-llm}             & 74.5 & 15.0 & 37.8 & 45.9 & 57.8 & 48.6 \\
OccLLaMA~\cite{occllama}               & 79.9 & 18.9 & 42.8 & 49.1 & 65.2 & 53.4 \\
LSceneLLM~\cite{lscenellm}             & 83.6 & 19.6 & 44.8 & 53.8 & 68.7 & 56.4 \\
BEVDet+BUTD~\cite{nuscenes-qa}         & 83.7 & 20.9 & 48.8 & 52.0 & 67.7 & 57.0 \\
CenterPoint+BUTD~\cite{nuscenes-qa}    & 84.1 & 21.3 & 49.2 & 55.9 & 69.2 & 58.1 \\
OpenDriveVLA~\cite{opendrivevla}       & 84.2 & 22.7 & 49.6 & 54.5 & 68.8 & 58.2 \\
\midrule
\multicolumn{7}{l}{\textbf{\textit{Zero-shot}}} \\
LLaMA-Adapter V2~\cite{llava-adapterv2} & 19.3 &  2.7 &  7.6 & 10.8 &  1.6 &  9.6 \\
LLaVA1.5~\cite{llava1-5}               & 45.8 &  7.7 &  7.8 &  9.0 & 52.1 & 26.2 \\
\textbf{Chat-Scene++ (Ours)} & \textbf{58.2} & \textbf{11.4} & \textbf{29.1} & \textbf{35.7} & \textbf{54.9} & \textbf{39.8} \\
\bottomrule
\end{tabular}}
\end{table}

\noindent \textbf{Effects of cross-modal object features.}  
Table~\ref{tab:encoder_ablate} shows results of various methods for both 3D and 2D feature extraction.  

For 3D feature extraction, we primarily compare three approaches: 1) Directly using the hidden object features from the Mask3D~\cite{mask3d} detector; 2) Extracting object-centric features using the pre-trained object-level encoder Uni3D~\cite{uni3d}; 3) Enhancing Mask3D features with the CLASP~\cite{grounded3dllm} pre-training method. The results in Rows 1, 2, and 5 indicate that the third approach is the most effective. Additionally, for Uni3D features, we experiment with incorporating position embeddings of each object's location to provide spatial context. However, the results in Row 3 suggest that position embeddings do not significantly improve performance.  

For 2D feature extraction, we compare two large-scale pre-trained 2D encoders: DINOv2~\cite{dinov2} and CLIP~\cite{CLIP}. The results in Rows 6 and 7 demonstrate that DINOv2 outperforms CLIP. This difference may stem from their distinct pre-training paradigms: CLIP, trained with an image-text contrastive objective, captures high-level semantics from captions but often overlooks fine-grained pixel details due to the limited granularity in text supervision. In contrast, DINOv2, trained with self-supervised objectives at both the image and patch levels, captures richer local object details such as shape and texture, which are crucial for object representations.  

\noindent \textbf{Ablations on LLM Backbone.}
We evaluate different sizes and types of LLM backbones, as shown in Table~\ref{tab:llm_ablate}. For the Vicuna series models, we assess various sizes, including 1B, 7B, and 13B. We use Vicuna-7B~\cite{chiang2023vicuna} in our main experiments, as it is a common choice in previous LLM-based methods. Maintaining the same LLM backbone allows for a more insightful comparison of different model architectures in the multi-modal domain.

\begin{figure}[t]
     \centering
    \subfloat[w/o G-CoT.]{%
      \includegraphics[width=0.5\linewidth]{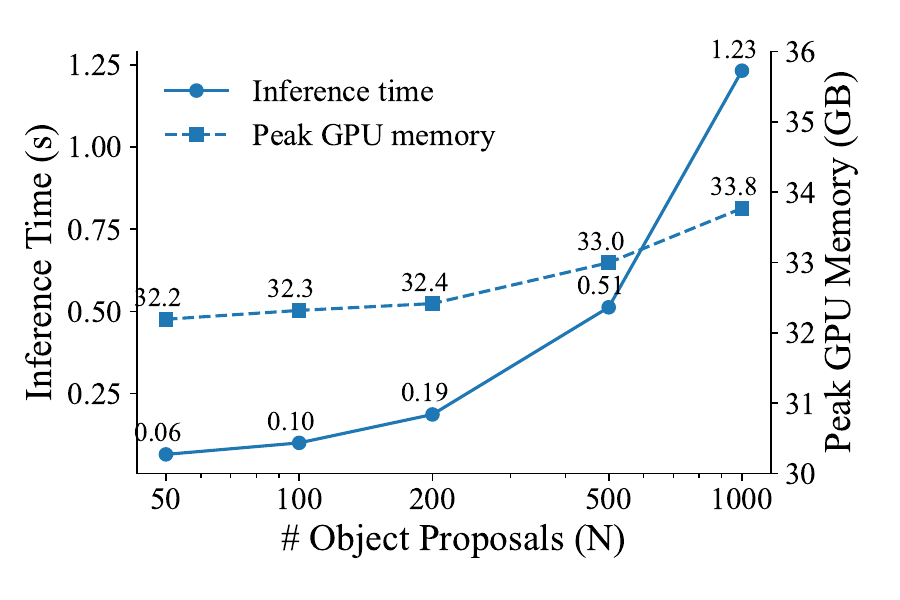}%
      \label{fig:latency_memory_no_gcot}%
    }
    \hfill
    \subfloat[w/ G-CoT.]{%
      \includegraphics[width=0.5\linewidth]{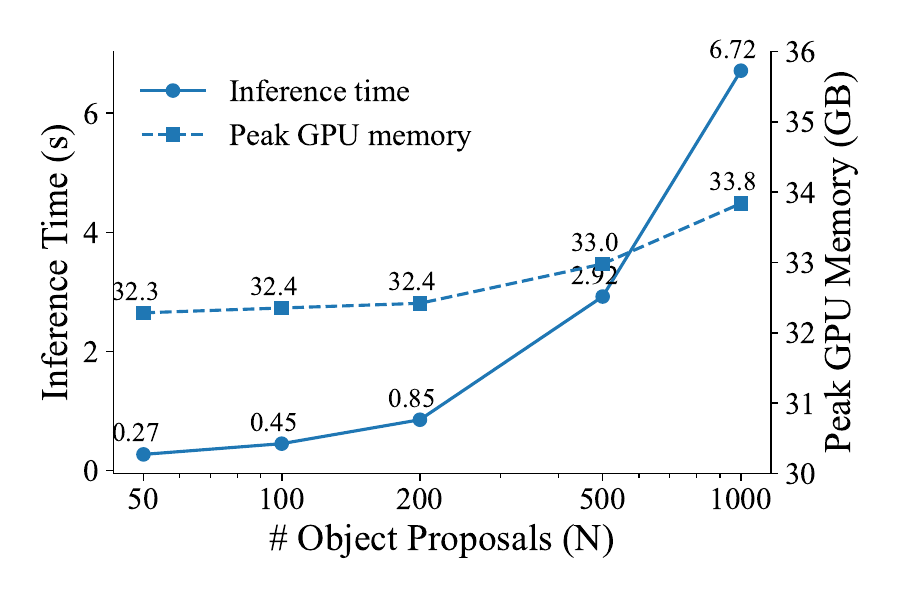}%
      \label{fig:latency_memory_with_gcot}%
    }
  
\caption{
    \revcommentsecond{New table}\addedsecond{\textbf{Scaling trend of inference time and GPU memory w.r.t.\ the number of object proposals.}
Chat-Scene++ uses 3 tokens per object and $N=100$ object proposals in main experiments. The inference time and GPU memory are measured during LLM forwarding on ScanQA.}
}
\label{fig:inference_time_memory_vs_N}
\end{figure}

\noindent\revcommentsecond{New experiment}\addedsecond{\textbf{Scaling of inference time and GPU memory w.r.t.\ the number of object proposals.}
Figure~\ref{fig:inference_time_memory_vs_N} reports both inference latency and peak GPU memory as a function of the number of object proposals $N$ (beyond $150$ up to $1000$). Since our representation uses a fixed budget of \emph{3 tokens per object}, the input length grows approximately linearly with $N$, leading to a monotonic increase in runtime. Importantly, the peak GPU memory remains nearly flat across this range, indicating that in our setup the additional proposals primarily affect compute time rather than causing prohibitive memory growth. Overall, these results suggest that Chat-Scene++ remains practical for moderate numbers of objects in large scenes, and that $N$ provides a simple knob to trade off efficiency and fidelity.}

\begin{figure}[tpb]
    \centering
    \includegraphics[width=\linewidth]{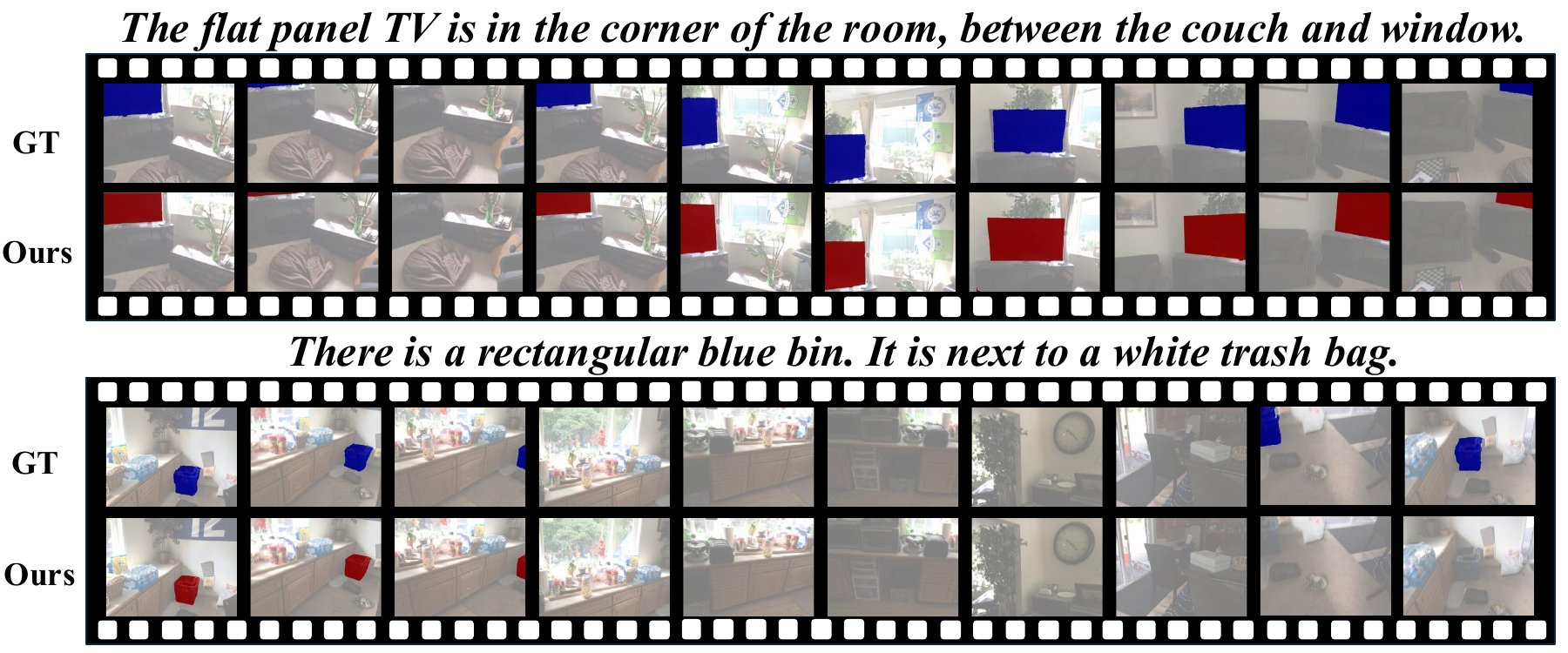}
    \caption{\textbf{Visualization results of visual grounding using 2D-only inputs (RGB scans).} ``GT'' denotes the 2D masks projected from the ground-truth 3D point cloud mask of the target object.}
    \label{fig:video_input_main}
\end{figure}

\noindent \textbf{Effects of grounded CoT.}  
We evaluate the effectiveness of grounded chain-of-thought (G-CoT) reasoning, as shown in Table~\ref{tab:cot_ablate}. The results in Rows 1 and 2 demonstrate that QA CoT significantly improves QA performance on ScanQA. This improvement can be attributed to the dataset’s annotations of semantically related objects, which provide explicit guidance for the model to focus on relevant context during reasoning. The results in Rows 1, 3, and 4 show that category-level CoT outperforms space-level CoT for grounding tasks in ScanRefer. This suggests that spatial proximity alone may not always provide the most useful context, as certain descriptions (e.g., ``farthest'') do not rely on nearby objects. Based on these findings, we adopt category-level CoT for grounding tasks in our main experiments. Combining both QA CoT and grounding CoT further yields consistent performance improvement, indicating their complementary benefits.

For the training schema, we compare one-stage and two-stage strategies. In one-stage training, all data is combined for joint learning, while in two-stage training, CoT data is excluded in the first stage and integrated in the second stage. Although two-stage training achieves a marginal improvement in grounding performance, it leads to a decline in QA performance. Considering this trade-off and efficiency, we adopt one-stage training for our experiments. Additional visualization results of grounded CoT are provided in the appendix.

Our analysis underscores the impact of semantically related object annotations in improving reasoning performance, as evidenced by significant gains on ScanQA. Extending this approach to grounding tasks like ScanRefer could further enhance the model’s ability to reason about object relationships and localize targets more accurately, presenting a promising direction for future 3D LLMs.

\subsection{Zero-shot Evaluation}

\noindent\revcommentsecond{New experiment}\addedsecond{\textbf{Zero-shot grounded CoT on SQA3D.}
Since we lack the necessary ``related objects'' training data for SQA3D, we are not able to create grounded CoT training data for this dataset. To directly test the generalization of our reasoning strategy, we perform a zero-shot G-CoT evaluation on SQA3D. As shown in Table~\ref{tab:sqa3d}, enabling G-CoT in a zero-shot setting improves the overall average performance. This demonstrates that our grounded reasoning strategy can generalize effectively to more challenging queries and new datasets even without direct fine-tuning, confirming the robustness of our approach. We further compute paired-bootstrap 95\% confidence intervals with 10{,}000 resamples over the test questions; while per-type gains can be mixed (e.g., slight drops on \emph{How} and \emph{Can}), the overall improvement remains stable under resampling.}

\noindent\revcommentfirst{New experiment}\addedfirst{\textbf{Open-vocabulary evaluation on 3RScan.} To test our model's ability to generalize to unseen categories and environments, we conduct a zero-shot evaluation on the 3RScan~\cite{3rscan} dataset using the 3RQA, 3RDialog, and 3RPlan tasks, following the protocol from LEO~\cite{leo}. Since our model was trained exclusively on ScanNet, this protocol rigorously tests generalization to both unseen object categories and novel environments. As shown in Table~\ref{tab:3rscan_results}, in the zero-shot setting, our model outperforms LEO across all three tasks, highlighting its superior ability to transfer knowledge to a new domain without direct training. Furthermore, after fine-tuning on the 3RScan dataset, our model extends its lead, achieving significant performance gains. This showcases both the robustness of our pre-trained model and its powerful capacity for adaptation.}

\noindent\revcommentsecond{New experiment}\addedsecond{\textbf{Zero-shot evaluation on NuScenes-QA~\cite{nuscenes-qa}.}
Table~\ref{tab:nuscene-qa} reports zero-shot transfer to nuScenes-QA. Chat-Scene++ achieves 39.8\% overall accuracy in the zero-shot setting, with consistent performance across all question types. We emphasize that our model is trained primarily on indoor 3D scene data, whereas nuScenes-QA targets outdoor autonomous-driving scenarios. This domain shift is substantial, spanning different environments, sensors, object distributions, and scene dynamics. Therefore, it is expected that methods finetuned on nuScenes-QA achieve higher absolute accuracy. Despite this large indoor-to-outdoor gap, the strong zero-shot performance of Chat-Scene++ indicates that our representation and grounded reasoning framework generalize well to autonomous-driving scenes without task-specific fine-tuning.}

\begin{table}[tbp]
\centering
\caption{\textbf{Evaluation results on ScanNet tasks using 2D-only input.} ``Random'' denotes predictions made using randomly selected instances, while ``Upperbound'' represents the best-matched instances from the segmented results.}  
 \label{tab:video_input}
\resizebox{0.8\linewidth}{!}{
\begin{tabular}{ccccc}
\toprule
\multirow{2}{*}{Method} & \multicolumn{2}{c}{Visual Grounding} & ScanQA & SQA3D \\
 & Acc@0.25 & Acc@0.5 & CIDEr & EM \\ \midrule
Random & 1.4 & 0.5 & -- & -- \\
Ours & \textbf{22.8} & \textbf{10.8} & \textbf{85.6} & \textbf{52.9} \\
{\color{gray}Upperbound} & {\color{gray}54.9} & {\color{gray}22.2} & {\color{gray}--} & {\color{gray}--} \\
\bottomrule
\end{tabular}}
\end{table}
% \end{wraptable}

\begin{table}[t]
\centering
\caption{\revcommentsecond{Revised table with stronger baselines LLaVA-Video-7B and InterVL2-8B}\addedsecond{\textbf{Performance comparison on VSI-Bench~\cite{vsi-bench} using 2D-only input.} ``Avg.'' denotes overall performance.}}
\label{tab:vsibench_results}

\resizebox{\linewidth}{!}{
\begin{tabular}{@{}cc cccc cccc@{}}
\toprule

& & \multicolumn{4}{c}{\textbf{Numerical Answer}} & \multicolumn{4}{c}{\textbf{Multiple-Choice Answer}} \\
\cmidrule(lr){3-6} \cmidrule(lr){7-10}

\textbf{Methods} & \textbf{Avg.} &
\rot{\textbf{Obj. Count}} &
\rot{\textbf{Abs. Dist.}} &
\rot{\textbf{Obj. Size}} &
\rot{\textbf{Room Size}} &
\rot{\textbf{Rel. Dist.}} &
\rot{\textbf{Rel. Dir.}} &
\rot{\textbf{Route Plan}} &
\rot{\textbf{Appr. Order}} \\
\midrule

LongVILA-8B & 21.6 & 29.1 & 9.1 & 16.7 & 0.0 & 29.6 & 30.7 & 32.5 & 25.5 \\
VILA-1.5-8B & 28.9 & 17.4 & 21.8 & \textbf{50.3} & 18.8 & 32.1 & 34.8 & 31.0 & 24.8 \\
LongVA-7B & 29.2 & 38.0 & 16.6 & 38.9 & 22.2 & 33.1 & \textbf{43.3} & 25.4 & 15.7 \\
LLaVA-OV-7B & 32.4 & 47.7 & 20.2 & 47.4 & 12.3 & 42.5 & 35.2 & 29.4 & 24.4 \\
LLaVA-Video-7B & 35.6 & 48.5 & 14.0 & 47.8 & 24.2 & \textbf{43.5} & 42.4 & \textbf{34.0} & 30.6 \\
InternVL2-8B & \textbf{37.5} & 31.3 & 29.0 & 48.9 & \textbf{44.2} & 38.0 & 33.4 & 28.9 & \textbf{46.4} \\
Chat-Scene++ & 32.4 & \textbf{58.3} & \textbf{30.9} & 20.3 & 21.5 & 29.7 & 40.6 & 21.9 & 36.2 \\
\bottomrule
\end{tabular}}
\end{table}

\subsection{Real-World Application with 2D-Only Inputs}
In practical applications of 3D scene understanding, capturing indoor RGB video scans is more straightforward than reconstructing 3D point clouds from RGB-D images. To evaluate our model's adaptability to 2D-only input (RGB scans) for 3D indoor scenes, we conduct experiments on ScanNet~\cite{scannet} and VSI-Bench~\cite{vsi-bench}.

Our process for video is as follows: First, we use the video detector DEVA~\cite{deva} to identify and track objects across frames. Next, we extract features for tracked objects using the Uni3D~\cite{uni3d} encoder. For visual grounding task, we evaluate the grounding results using 2D masks on the video frames.

\noindent \textbf{Tasks and metrics.}
For ScanNet dataset, we evaluate visual grounding and VQA on 2D video. For visual grounding, we project the ground-truth 3D object onto 2D masks in the video frames to calculate the Intersection over Union (IoU). To evaluate performance over an entire video of length $L$, we concatenate the predicted per-frame masks $\{\mathbf{M}^{\mathrm{p}}_i\in\mathbb{R}^{H \times W} \}_{i=1...L}$ and ground-truth masks $\{\mathbf{M}^{\mathrm{g}}_i\in\mathbb{R}^{H \times W} \}_{i=1...L}$ into spatiotemporal volumes, $\mathbf{\tilde{M}}^\mathrm{p}$ and $\mathbf{\tilde{M}}^\mathrm{g}$, respectively. We then propose a Spatial-Temporal IoU (ST-IoU) to measure the overlap between these volumes. Using this metric, we report accuracy at 0.25 and 0.5 ST-IoU thresholds (Acc@0.25, Acc@0.5). For VQA, we use the standard annotations and metrics from ScanQA and SQA3D.

\noindent\revcommentfirst{New experiment}\addedfirst{For VSI-Bench, we simply follow their standard evaluation tasks and protocols: mean relative accuracy for numerical answers, and accuracy for multiple-choice answers.}

% \noindent\textbf{Performance Analysis.} For ScanNet tasks, as shown in Table~\ref{tab:video_input}, our method achieves strong grounding performance, significantly outperforming a random baseline and approaching the upper bound set by the video detector's mask quality. A key limitation, visualized in Figure~\ref{fig:video_input_main}, is that the video detector may lose track of objects that go out of view, reducing the quality of the extracted masks and lowering the upper-bound Acc@0.5 score. For VQA, our results are comparable to those using 3D inputs. This demonstrates that our approach effectively builds a robust scene understanding from sequences of object embeddings, even when the object proposals are imperfect.

\noindent\revcommentsecond{Revised analysis}\addedsecond{\textbf{Performance Analysis.} For ScanNet tasks, as shown in Table~\ref{tab:video_input}, our method achieves strong grounding performance, significantly outperforming a random baseline and approaching the upper bound set by the video detector's mask quality. A key limitation, visualized in Figure~\ref{fig:video_input_main}, is that the video detector may lose track of objects that go out of view, causing temporal drift/identity switches and reducing the precision of the extracted masks, which in turn lowers the upper-bound Acc@0.5 score. This also explains the observed discrepancy between 2D-only VQA and localization metrics: while IoU-based localization (e.g., Acc@0.25/0.5) is highly sensitive to mask boundary accuracy and spatial alignment, VQA (e.g., EM) is often more tolerant to moderate proposal noise as long as the correct semantic instance is roughly identified. Consequently, our 2D-only setting can achieve VQA performance comparable to 3D inputs, yet remain constrained on visual localization by the upstream video detector. Overall, these results demonstrate that our approach can build robust scene understanding from sequences of object embeddings even with imperfect proposals, and that improving video detection/tracking quality would directly translate into higher 2D-only localization performance without changing the reasoning framework.}

\revcommentsecond{Revised analysis}\addedsecond{For VSI-Bench, the results in Table~\ref{tab:vsibench_results} indicate that Chat-Scene++ achieves competitive performance under the 2D-only input setting. In particular, it demonstrates strong numerical, object-centric reasoning ability, achieving the best results on \textit{Obj. Count} and \textit{Abs. Dist.}, which reflects effective quantitative spatial understanding enabled by structured object representations. We also observe that some 2D MLLM baselines (e.g., InternVL2-8B and LLaVA-Video-7B) obtain stronger overall scores primarily because they are pre-trained on substantially larger-scale 2D multimodal corpora (often millions of image/video--text pairs) with broad VQA-style supervision that aligns well with VSI-Bench. In contrast, Chat-Scene++ is trained exclusively on 3D scene datasets with a much smaller scale (approximately 300K samples) and thus benefits less from diverse 2D VQA pretraining. Despite this modality and data-scale gap, the competitive results support our claim that representing scenes as structured object sequences enhances spatial reasoning and generalizes well to 2D-only tasks.}

\section*{Acknowledgments}
This work was done during an internship at Shanghai AI Laboratory.
% This work was supported by National Natural Science Foundation of China under Grant No.U24A20326 and National Natural Science Foundation of China under Grant No.62572423.

\let\oldthebibliography\thebibliography
\let\endoldthebibliography\endthebibliography

\renewenvironment{thebibliography}[1]
  {\oldthebibliography{#1}\fontsize{7.8pt}{9pt}\selectfont}
  {\endoldthebibliography}

% \newpage
{
\bibliographystyle{IEEEtran}
\bibliography{main}
}

\section{Biography Section}
\vspace{-33pt}
\begin{IEEEbiography}[{\includegraphics[width=1in,height=1.25in,clip,keepaspectratio]{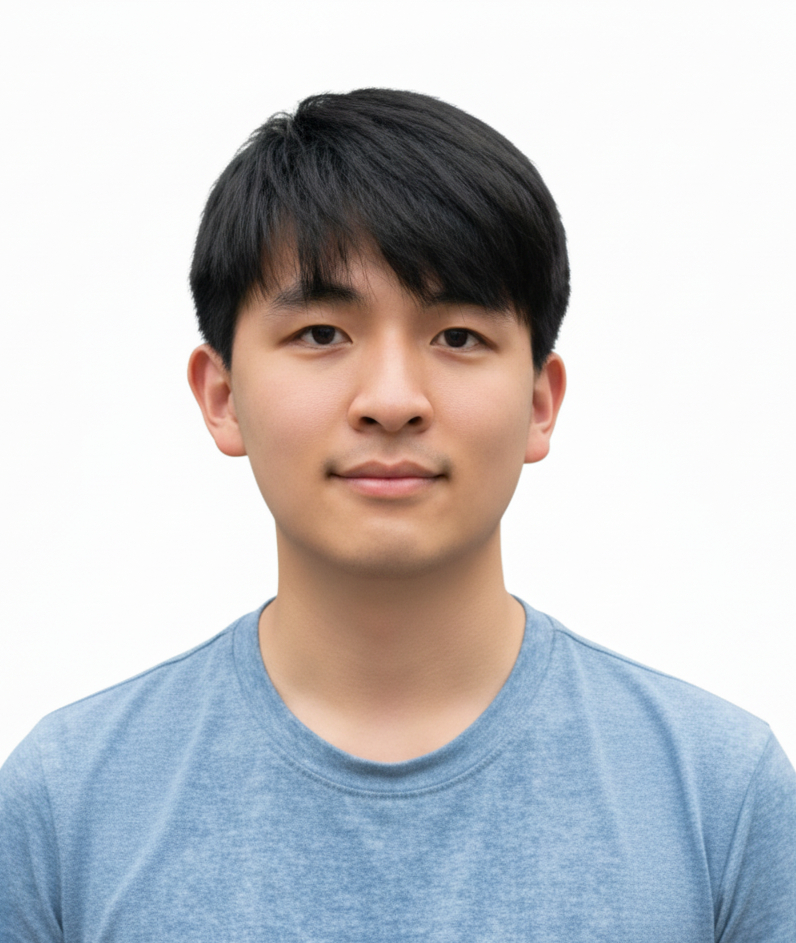}}]{Haifeng Huang}
received the BEng degree in Computer Science from the College of Computer Science and Technology, Zhejiang University (ZJU) in 2022. He is currently working toward the MS degree with the Zhejiang University (ZJU), under the supervision of Prof. Zhou Zhao. He serves as a reviewer for NeurIPS, ICLR, ICML, CVPR, ICCV and etc. His research interests lie in computer vision, multi-modal learning, and 3D scene understanding.
\end{IEEEbiography}

\vspace{-33pt}
\begin{IEEEbiography}[{\includegraphics[width=1in,height=1.25in,clip,keepaspectratio]{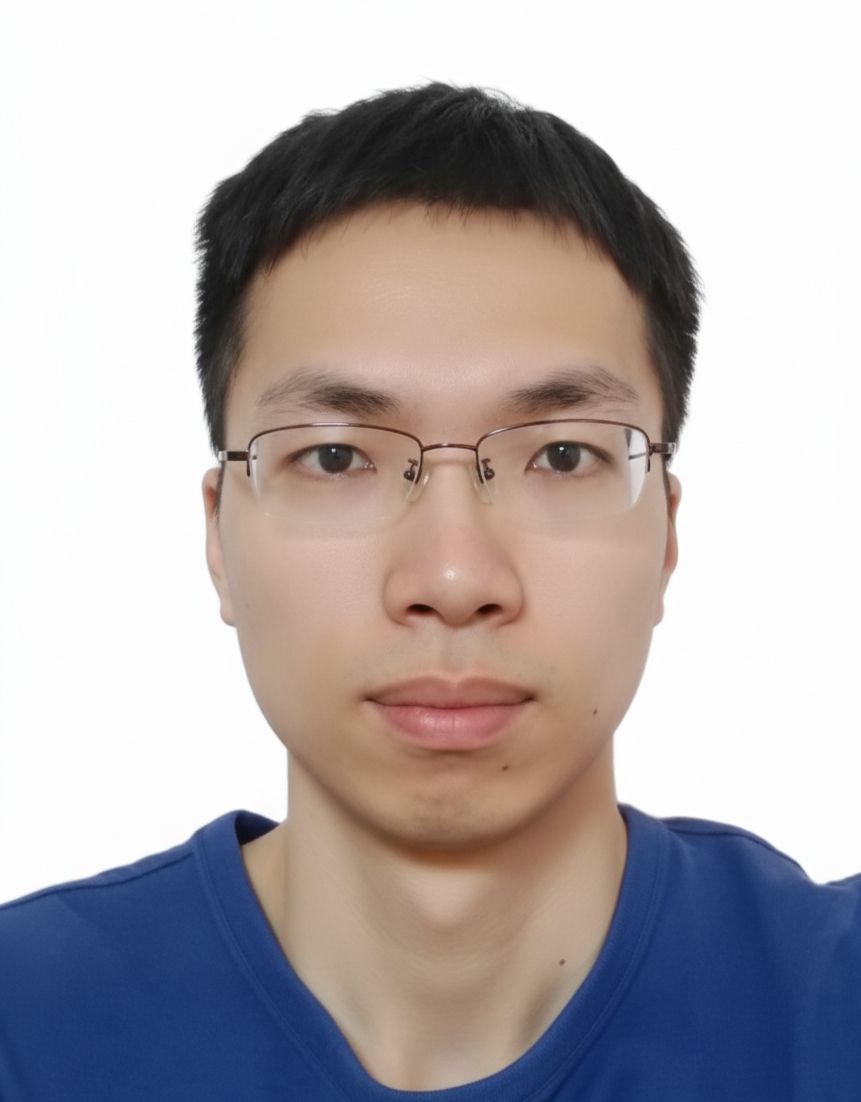}}]{Yilun Chen}
(Member, IEEE) received his Ph.D. from The Chinese University of Hong Kong, where he was supervised by Prof. Jiaya Jia, and his B.Eng. in Computer Science and Technology from Beihang University, China. He serves as a reviewer for T-PAMI, NeurIPS, CVPR, ICLR and etc. He is currently a researcher at the Shanghai AI Laboratory, focusing on 3D perception and embodied AI systems.
\end{IEEEbiography}

\vspace{-33pt}
\begin{IEEEbiography}[{\includegraphics[width=1in,height=1.25in,clip,keepaspectratio]{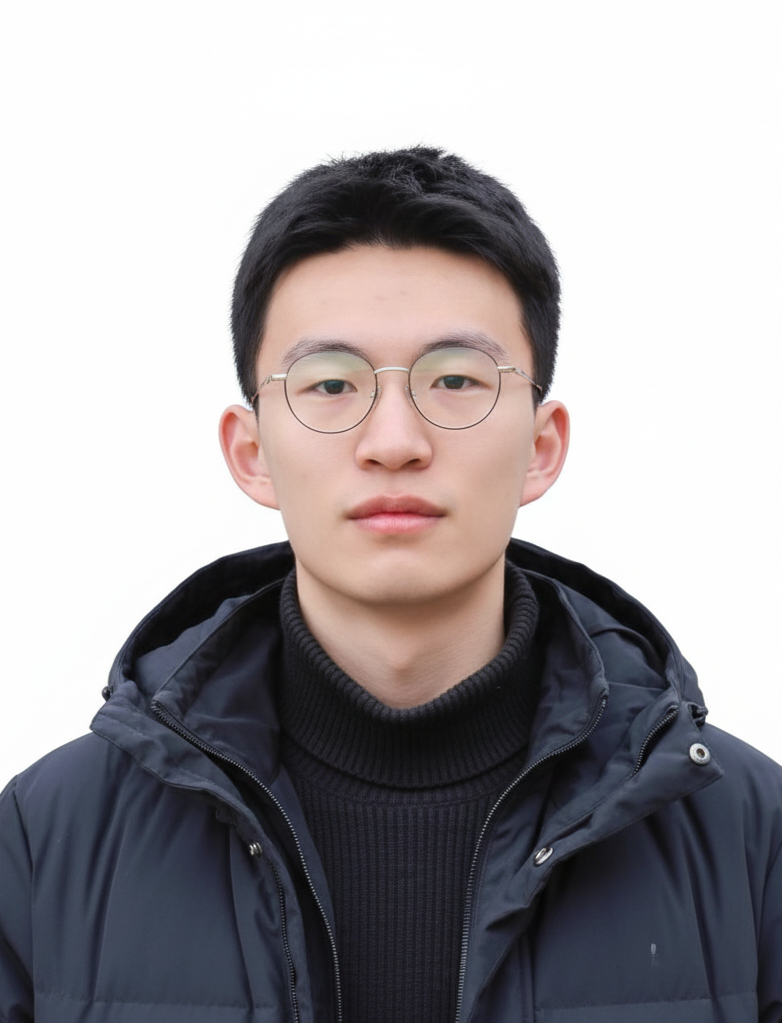}}]{Zehan Wang}
received the BEng degree in Biosystems Engineering from the College of Biosystems Engineering and Food Science, Zhejiang University (ZJU) in 2022. He is currently working toward the PhD degree with the Zhejiang University (ZJU), under the supervision of Prof. Zhou Zhao. He serves as a reviewer for NeurIPS, ICLR, ICML, CVPR, ICCV and etc. His research interests lie in spatial intelligence, multi-modal understanding and generation.
\end{IEEEbiography}

\vspace{-33pt}
\begin{IEEEbiography}[{\includegraphics[width=1in,height=1.25in,clip,keepaspectratio]{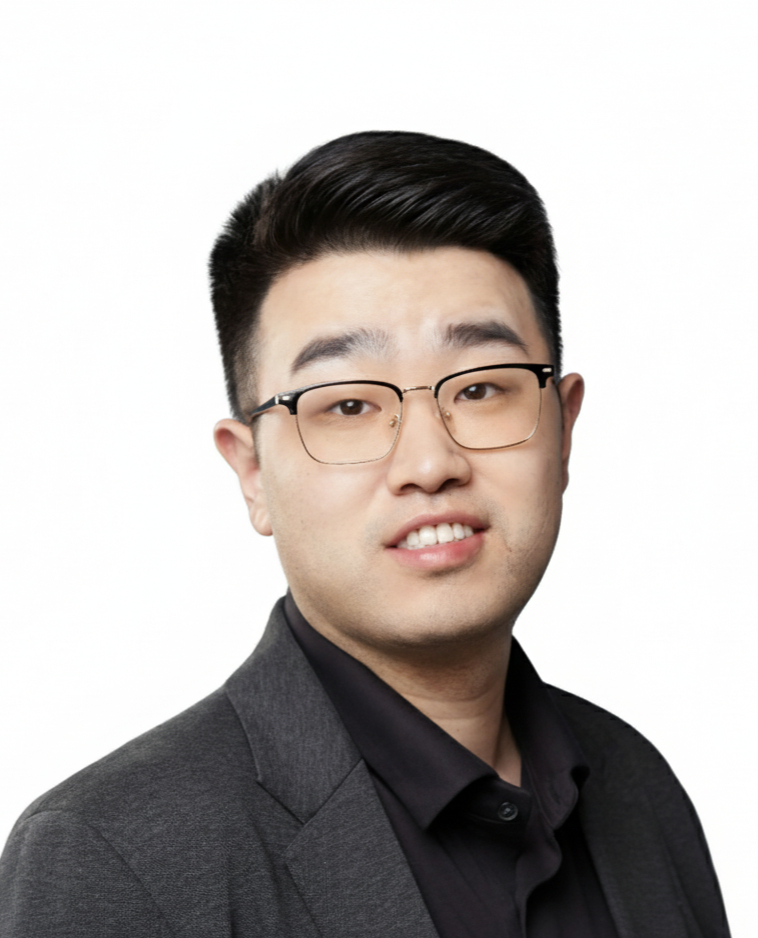}}]{Jiangmiao Pang}
is currently a Research Scientist at Shanghai AI Laboratory. His research interests span multi-modal learning and robotics. Recently, he has been dedicated to building foundation models and digital platforms for Embodied AGI. He has also led/authored some famous open-source projects, including MMDetection, MMTracking, and MMDetection3D.
\end{IEEEbiography}

\vspace{-33pt}
\begin{IEEEbiography}[{\includegraphics[width=1in,height=1.25in,clip,keepaspectratio]{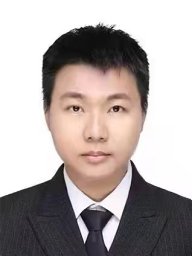}}]{Zhou Zhao}
(Member, IEEE) received the B.S. and Ph.D. degrees in Computer Science from The Hong Kong University of Science and Technology, Hong Kong, in 2010 and 2015, respectively. He is currently a Professor with the College of Computer Science and Technology, Zhejiang University. His research interests include machine learning and data mining.
\end{IEEEbiography}

\end{document}